\documentclass[10pt,twocolumn,letterpaper]{article}
\usepackage[pagenumbers]{cvpr} 
\usepackage[dvipsnames]{xcolor}

\usepackage{lipsum}
\usepackage{xstring}
\usepackage{placeins}
\usepackage{cuted}
\usepackage{float}
\usepackage{multirow}
\usepackage{array}
\usepackage{nicefrac}

\definecolor{cvprblue}{rgb}{0.21,0.49,0.74}
\usepackage[pagebackref,breaklinks,colorlinks,citecolor=cvprblue]{hyperref}

\title{Online Video Depth Anything:\\ 
Temporally-Consistent Depth Prediction with Low Memory Consumption}

\author{
Johann-Friedrich Feiden \quad
Tim Küchler \quad
Denis Zavadski \quad
Bogdan Savchynskyy \quad
Carsten Rother \\
Heidelberg University \\
{\tt\small \{firstname.lastname\}@iwr.uni-heidelberg.de}
}

\begin{document}
\maketitle

\begin{strip}
  \begin{minipage}[t]{0.63\linewidth}
    \centering
    \includegraphics[width=\linewidth]{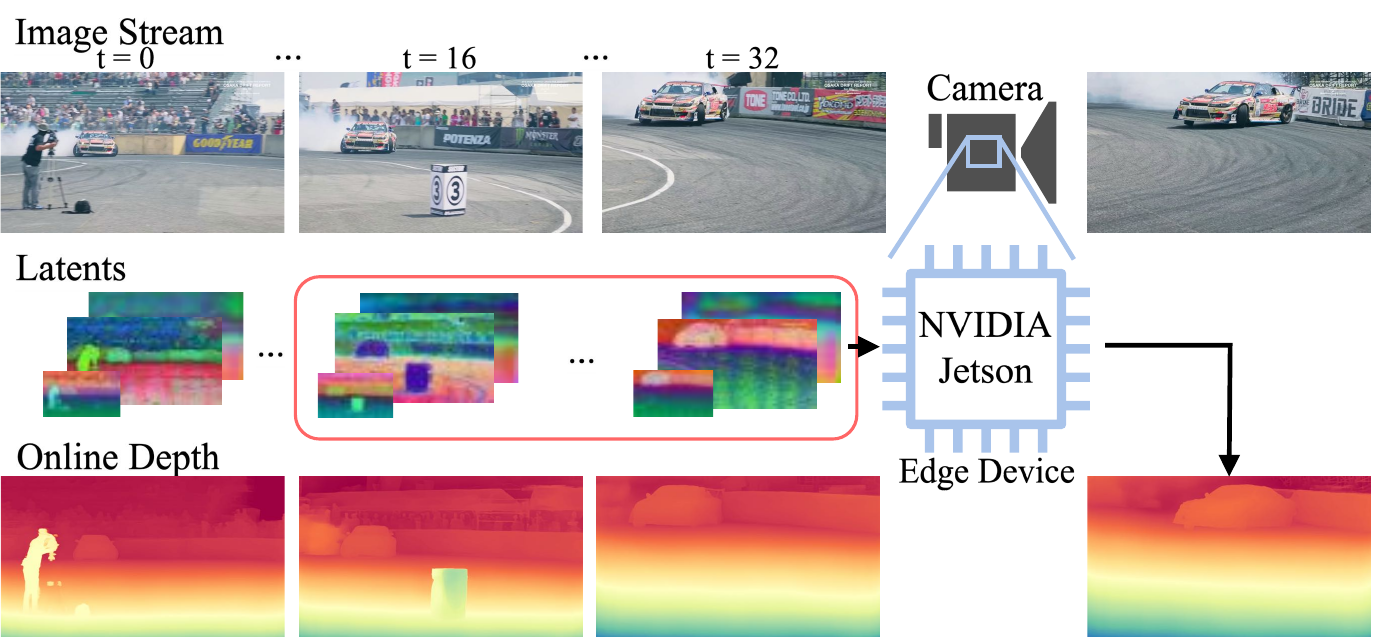}
  \end{minipage}
  \hfill
  \begin{minipage}[t]{0.31\linewidth}
    \centering
    \includegraphics[width=\linewidth, trim=2.5mm 3mm 2.5mm 2.5mm,
    clip]{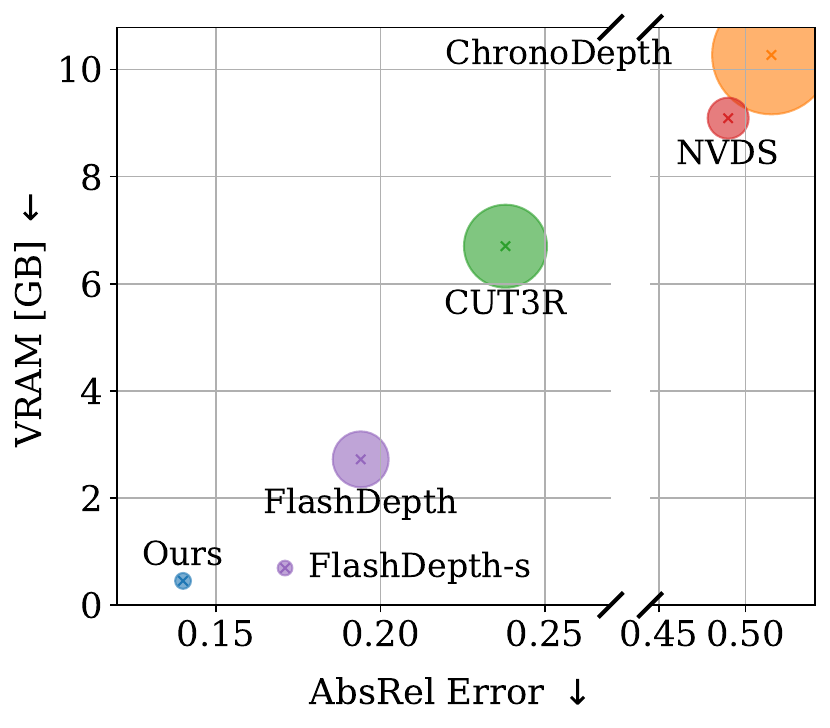}
  \end{minipage}
  \captionof{figure}{(Left) Our \textit{online Video Depth Anything} approach (oVDA) produces high-quality depth predictions at up to 20 FPS on an edge device (NVIDIA Jetson Orin NX) by caching a sliding window of past latent features (red box). (Right) Performance comparison of various \textit{online} video depth estimation methods for the KITTI \cite{Kitti} dataset. Our oVDA approach outperforms all competitors in both AbsRel error and VRAM usage. Note that low VRAM usage is crucial for many real-world applications, such as deployment on edge devices. The size of each circle represents the number of parameters. }
  \label{fig:MoneyPlot Vram}
  
\end{strip}
\begin{abstract}
\setlength{\parindent}{0pt}
Depth estimation from monocular video has become a key component of many real-world computer vision systems. Recently, Video Depth Anything (VDA) has demonstrated strong performance on long video sequences. However, it relies on batch-processing which prohibits its use in an online setting. In this work, we overcome this limitation and introduce \mbox{online VDA} (oVDA). The key innovation is to employ techniques from Large Language Models (LLMs), namely, caching latent features during inference and masking frames at training. Our oVDA method outperforms all competing online video depth estimation methods in both accuracy and VRAM usage. Low VRAM usage is particularly important for deployment on edge devices. We demonstrate that oVDA runs at 42 FPS on an NVIDIA A100 and at 20 FPS on an NVIDIA Jetson edge device. We will release both, code and compilation scripts, making oVDA easy to deploy on low-power hardware.
\end{abstract}
      
\section{Introduction}
\label{sec:intro}
Online monocular depth prediction has become an increasingly relevant topic for real-time, industrial applications, such as robotics, drone technology or augmented reality.  
This is primarily due to its potential to replace expensive sensor systems such as LiDAR with low-cost monocular cameras~\cite{DeepLearningBasedMonoDepth-Review}. 
We aim at developing a monocular video depth estimation method that can be applied to any input (zero-shot capability) and is fast and accurate while running on low-powered, low-memory hardware (see \cref{fig:MoneyPlot Vram}).   

The field of \textit{single-image} depth prediction has made tremendous strides forward in recent years~\cite{FirstDepth,depth_anything_v2,depthanything,OriginalMidas,Marigold,DepthPro,PrimeDepth}. Today, we have powerful foundation models for non-metric, i.e. scale- and shift-invariant, depth estimation working well across diverse domains.
Among these, Depth Anything v2~\cite{depth_anything_v2} stands out for its impressive performance and manageable computational cost, particularly in its smaller variant. 
However, applying such a non-metric, single-image method to each frame of a video sequence, naturally produces temporally flickering depth maps, necessitating frame-by-frame alignment or post-processing.

For \textit{video} depth estimation, the temporal dimension can be used to improve the performance with respect to single-image methods. But, at the same time, for many applications such as augmented reality, the temporal dimension poses the additional challenge 
of producing \textit{temporally consistent} depth maps. Earlier approaches to this challenge often relied on test-time training~\cite{RobustConsistentVideoDepthEstimation,VideoDepth-2020,ConDepthMovingObjects}. While this strategy is effective, it is computationally expensive and too slow for real-time settings~\cite{VDA}. 
Modern research instead focuses on feed-forward methods, which can be broadly grouped into three categories. However, methods from two categories do not have the potential to be deployed on low-powered hardware due to high computational cost and memory consumption, i.e. diffusion-based approaches~\cite{ChronoDepth,DepthCrafter,DepthAnyVideo} as well as methods that maintain an underlying 3D representation \cite{CUT3R,DUST3R,MONST3R,MAST3R,DroidSlam,BTimer,PointBasedFusion}. 
In contrast, methods of the third category that extend existing single-image  models~\cite{VDA,flashdepth,NVDS,RollngDepth,depthsync} do have this potential.

Our work falls into this third category. 
We build upon Video Depth Anything (VDA)~\cite{VDA}, a recent method that achieves state-of-the-art results for non-metric depth prediction of long videos. 
However, VDA is designed for \textit{offline} video processing and lacks support for online inference, as it can only process entire \textit{batches of frames} at a time.  
To address this limitation, we introduce oVDA, which transforms VDA to an online setting for predicting temporally consistent video depth maps. 
We draw inspiration from the inference strategies employed in Large Language Models (LLMs). In LLMs, it is common to cache previously computed latent features (i.e. a context window), allowing for efficient prediction of new tokens without recomputing the entire context~\cite{TransformersLibary,EfficientTransformerInference,MiniKVCache}. 
Analogously, oVDA stores latent features of past video frames and uses them as temporal context for predicting future depth maps. 
To efficiently train our model, we again draw inspiration from LLMs. 
We adapt masked attention~\cite{AttentionIsAllYouNeed}, which enables the training process to selectively attend to past frames while preventing the model from using future frames.
Finally, to further enhance temporal consistency we introduce a new scale-and-shift consistency loss.
Our contributions are:
\begin{itemize}
    \item We introduce oVDA which transforms VDA to an online depth prediction method by employing LLM-inspired inference and training techniques, as well as a new scale-and-shift consistency loss.  
    \item oVDA outperforms competing online video depth estimation methods, both in terms of accuracy and VRAM usage.
    It runs at 42 FPS on an NVIDIA A100 and at 20 FPS on an NVIDIA Jetson edge device. 
    \item We will release both, code and the compilation script for oVDA, making it easy to deploy on low-power edge device hardware such as NVIDIA Jetson.
\end{itemize}

\section{Related Work}
\label{sec:RelatedWork}
Early learning-based, single-image depth estimation methods were limited to one datasets and constrained scenarios~\cite{FirstDepth,DeepConvForDepth,OldSingleImageDepthOverview}. 
The introduction of the scale- and shift-invariant loss by MiDaS~\cite{OriginalMidas} redirected the focus towards multi-dataset training and zero-shot evaluation. 
With powerful backbones such as Stable Diffusion~\cite{ldm} or DINOv2~\cite{DINOv2} and access to vast amounts of training data, numerous zero-shot single-image depth estimation approaches have emerged~\cite{depth_anything_v2,depthanything,Marigold,PrimeDepth,D3vo,Geowizard}, capable of predicting scale- and shift-invariant depth for almost any image. 
While these methods achieve outstanding results for single-images, they produce flickering and temporally inconsistent depth maps when applied frame-by-frame to videos~\cite{NVDS}. Note that \textit{metric}, single-image depth prediction has been less popular due to limited amount of ground truth data.  

{\bf Monocular Video Depth Estimation} aims at exploiting temporal information in order to improve on single-image methods and, at the same time, reduce flickering artefacts. Early works relied on test-time training~\cite{VideoDepth-2020,RobustConsistentVideoDepthEstimation,ConDepthMovingObjects}, for instance by using photometric or geometric constraints over multiple frames.  However, due to their high computational cost, such methods are rarely suitable for real-time inference and often have to operate at low resolutions~\cite{VideoDepth-2020,VDA}.
Using LSTMs~\cite{LSTM} for temporal information flow has also been explored~\cite{DontForgetPast,ExploitingTemporalConsis,OnlineDynamicDepth_2023}, but since these methods are typically trained on relatively small datasets, their zero-shot capability is limited. 
Moreover, despite recurrent blocks, they still tend to produce flickering results~\cite{NVDS}.

Modern approaches follow one of three strategies.
The first leverages video diffusion models~\cite{StableVideoDiffusionDatasets,AlignLatents}, which possess a strong understanding of temporal dynamics. 
Methods such as~\cite{ChronoDepth, DepthCrafter, DepthAnyVideo} adapt these models to predict depth, resulting in highly consistent temporal and spatial predictions. This, however, comes at the cost of high computational demand, or, in some methods, batch-wise processing~\cite{DepthCrafter, DepthAnyVideo}, which prevents their use in online settings.

The second strategy stores a continuous 3D representation for a video. DUSt3R~\cite{DUST3R} estimates a point cloud from an unconstrained image collection, from which camera parameters and depth can be derived. 
Building upon this idea, several works~\cite{CUT3R,MONST3R,MAST3R,DroidSlam,BTimer,PointBasedFusion} predict depth for multiple frames and subsequently perform global alignment, which again hinders online usage. Addressing this, other methods~\cite{CUT3R, streamVGGT} construct a persistent state that is updated with each incoming frame, resulting in long, consistent video predictions.

In the third strategy, single-frame methods are extended with temporal modules, thereby preserving the strong single-image prior. 
For example, NVDS~\cite{NVDS} utilises an off-the-shelf single-image depth predictor, followed by a smoothing network that temporally aligns the predictions. 
While this allows for online predictions, best results still require backward refinement which is not available in an online setting. 
Another approach is to perform temporal alignment directly within the decoder. 
Video Depth Anything (VDA)~\cite{VDA}, for instance, uses Depth Anything v2 (DAv2) ~\cite{depth_anything_v2} as its image backbone, achieving results surpassing DAv2. Due to training and architectural constraints, VDA can only process videos in batches, limiting its applicability to online depth estimation. 
FlashDepth~\cite{flashdepth} also builds upon DAv2 and extends it with the state-space model MambaV2~\cite{Mamba1,Mamba2} for incorporating temporal information more efficiently, while at the same time enabling online prediction.
Other learning-based approaches, such as MAMo~\cite{MAMo}, introduce a temporal memory which is updated during inference. However, such an approach cannot modify the backbone predictions, leading to flickering depth for longer videos~\cite{NVDS}. 

Our work follows the third strategy. We transform VDA to an online  method, which we denote as oVDA, that runs at 42 FPS on an NVIDIA A100. We compare against ChronoDepth~\cite{ChronoDepth}, NVDS~\cite{NVDS}, FlashDepth~\cite{flashdepth} and CUT3R~\cite{CUT3R} as recent online methods, and show that we outperform them in terms of accuracy and VRAM usage.

\section{Method}
\label{sec:Method}
In \cref{subsec:Architecture} we present the architecture of the Video Depth Anything (VDA) model, on which we build. For our online Video Depth Anything (oVDA) method, we improve on three aspects of VDA. Firstly,     
\cref{subsec:Inference} describes our online inference strategy. Secondly, \cref{subsec:Training} explains our new training procedure, and finally, \cref{subsec:Loss} introduces the new loss function of oVDA.  

\begin{figure*}[ht]
  \centering
   \includegraphics[width=\linewidth]{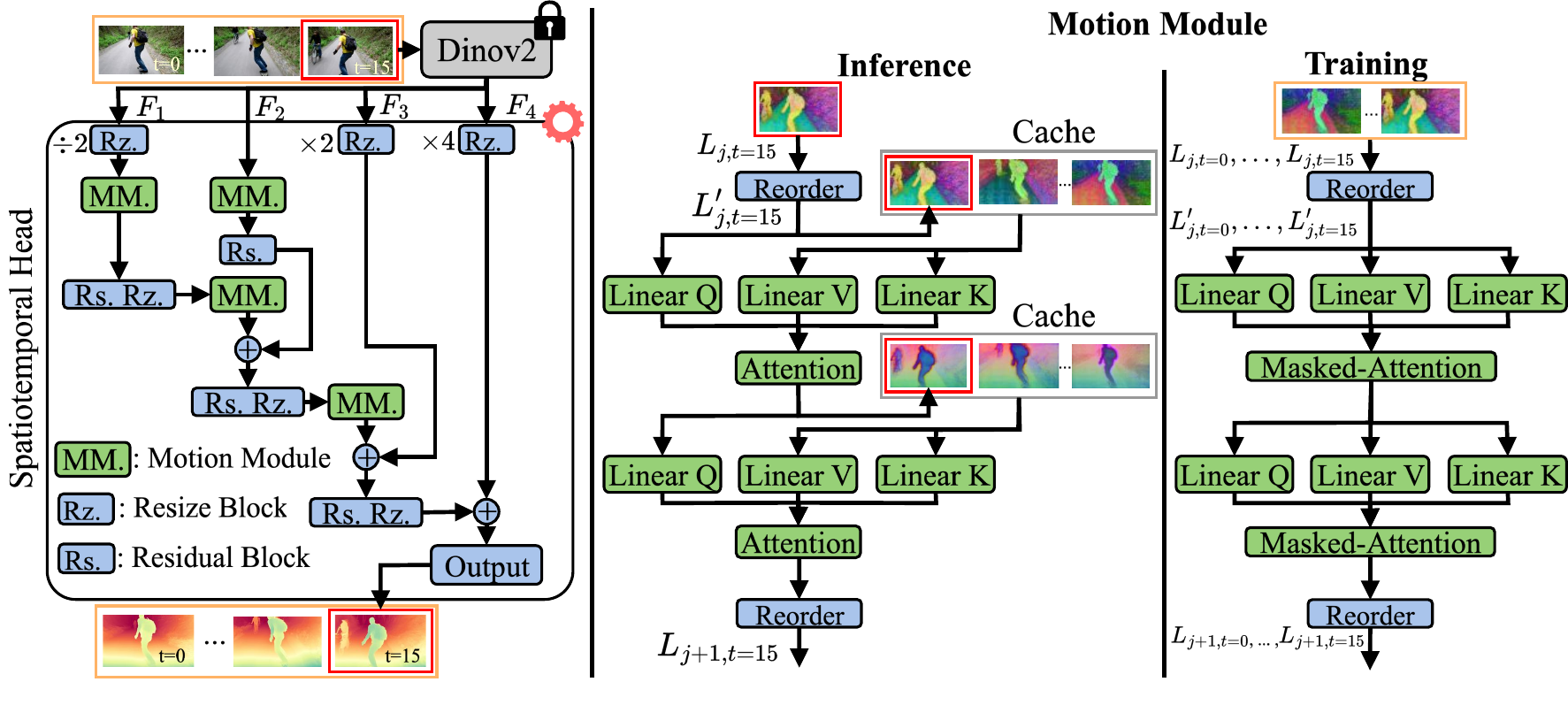}

   \caption{(Left) Our oVDA Architecture operating on the current frame $t=15$. In contrast to offline VDA, which runs in batches of frames (orange box), our online oVDA approach processes only the current frame (red box). The blue blocks process only spatial dimensions, while the Motion Modules (green) perform temporal reasoning and differ from VDA. During training, only the Spatiotemporal Head is fine-tuned, while we keep the DINOv2 backbone frozen. (Middle and Right) Illustration of the inference and training procedure of our new Motion Module with number $j$. Firstly, the hidden features are re-ordered to perform temporal reasoning. During inference, the cache is updated by adding the current latent feature ($L_{j,t=15}^\prime$) and removing the latent feature of the temporally last frame. Afterwards, cross-attention is applied between the current latent feature and the cached latent features. At training time, similar to VDA, we process a batch (here $t=0,...,15$) of frames but apply masked self-attention, ensuring that the current frame can only attend to past frames.}
    \vspace{-10px}
   \label{fig:Architecture}
\end{figure*}

\subsection{VDA Architecture}
\label{subsec:Architecture}
VDA has demonstrated strong performance for offline, non-metric depth prediction of very long video sequences~\cite{VDA}. 
Our oVDA method uses the same basic architecture as VDA, as well as its pre-trained weights. For computational efficiency we build upon VDA-s(mall). The key difference between our oVDA method and VDA is that VDA processes frames in parallel, i.e., in batches, whereas oVDA processes only a single (the current) frame at a time. This change requires to redesign the VDA Motion Modules (see \cref{subsec:Inference}), since these are the only blocks that perform temporal reasoning, while we keep all other VDA blocks unchanged. Our full oVDA architecture is shown in \cref{fig:Architecture} (left).

VDA uses a DINOv2 encoder for all frames of a batch, inspired by single-image Depth Anything v2 (DAv2)~\cite{depth_anything_v2}.
Given input images $X \in \mathbb{R}^{N \times C \times H \times W}$, where $N$ is the number of frames, $C$ the number of channels, and $H$, $W$ the height and width, the encoder extracts feature maps $F_i \in \mathbb{R}^{N \times \frac{H}{p} \times \frac{W}{p} \times C_i}$. 
Here, $p$ denotes the patch size, and $i$ is the index of the transformer block from which the feature map originates. 
For clarity, we ignore the batch dimension in our notation, as it can be incorporated into the number of frames.
Since DINOv2 has no temporal information flow, the extracted features are independent for each frame.

The extracted \mbox{DINOv2} features are first resized by the Spatiotemporal Head according to their layer within the encoder, with factors $4, 2, 1$ and $1/2$ respectively. 
These latent features $L_{j}$ are then transformed and iteratively fused by Motion Modules, Residual Blocks, and Resize Blocks (see \cref{fig:Architecture} (left)). 
Here $j$ is the block number (in total 12 blocks) within the Spatiotemporal Head. To enable temporally consistent predictions, VDA uses temporal self-attention within the Motion Modules. 
For this, the temporal and spatial dimensions are reordered, allowing attentions to operate across time (frames):
$Att(L_{j}’)$ with $L_j^\prime \in \mathbb{R}^{(\frac{H}{p} \times \frac{W}{p}) \times N \times C_i}$.
At the end of a Motion Module, the original tensor shape is restored, and the head continues processing in a frame-wise manner.
As shown in VDA~\cite{VDA}, the Motion Modules are powerful enough to maintain temporal consistency while even improving depth prediction quality compared to the single-frame DAv2 model. 

To transform the batch-wise VDA method to an online setting, where only the depth of the current frame is predicted, a self-attention mechanism is no longer applicable. The reason is that future frames are not yet available, and, secondly, depth predictions of past frames are fixed. 

\subsection{oVDA Inference Strategy}
\label{subsec:Inference}
To enable online depth prediction for VDA, we draw inspiration from Large Language Models (LLMs). 
Just as LLMs predict the next token given a sequence of previous tokens~\cite{AttentionIsAllYouNeed,TransformersLibary}, our goal is to predict the depth of the current frame using a moving window of the past $c$ frames as context. 
Analogously to LLMs that cache previously hidden states, such that predicting a new token does not require recomputation of the whole context window, we also cache the hidden representations of past frames and reuse them during inference to predict the next frame. 
Instead of performing self-attention across all frames in the context window (e.g. 32 for VDA), our approach reduces to cross-attention at inference time, using the cache as attention context (i.e. keys $K$ and values $V$) for predicting the current frame. This change allows for a more efficient forward pass, leveraging past information at real-time instead of batch-processing. Although this approach does work without retraining the model, it results in a noticeable drop in performance compared to VDA (see \cref{subsec:SaSCon}). We overcome this performance drop with our LLM-inspired training technique (see \cref{subsec:Training}).

As shown in \cref{fig:Architecture} (middle), we cache hidden features $L_{j,t-c}^{\prime},\dots,L_{j,t}^{\prime}$ before every attention within each Motion Module, where $t$ is the index of the current frame and $c$ the maximum context length.
At the beginning of a sequence, when no prior context is yet available, we apply standard self-attention. As subsequent frames arrive, the context window is incrementally populated until it reaches its maximum context length $c$. Beyond this point, the context operates as a sliding window over the video sequence, while always discarding the temporally last cached frame to maintain a fixed-size context window.

Unlike LLMs, which typically employ context windows spanning tens of thousands of tokens, we limit our context length to $c = 16$ for efficiency (see ablation in supplement).

\subsection{oVDA Fine-Tuning}
\label{subsec:Training}
We propose to use an LLM-inspired training approach called \textit{masked attention} (see \cref{fig:Architecture} (right)). 
In LLMs, masked attention forces the model to predict the next token as a function of previous tokens by masking the attention weights as a lower-triangular matrix, while still allowing for batch-wise propagation through the network during training~\cite{AttentionIsAllYouNeed}. 
Analogously, we apply attention masking within the Motion Modules of the Spatiotemporal Head, ensuring that past frames influence future frames, but not vice versa. This design enables batch-wise training to preserve a training procedure consistent with that of VDA.

Since we utilise pre-trained VDA weights, which were already optimised for video depth prediction, we only need to fine-tune the model for the online setting.
Taking this into account, we train with a comparably small learning rate of $10^{-6}$ ($10^{-4}$ for VDA) and a batch size of 16 for 130,000 steps, using a cosine learning rate scheduler.

To enable dense supervision and given that VDA has already been trained on real-world data, we fine-tune exclusively on synthetic datasets. 
Specifically, we use IRS~\cite{IRS}, PointOdyssey~\cite{PointOdyssey}, where we filter PointOdyssey scenes that lack background depth, TartanAir~\cite{TartanAir} and VKITTI~\cite{VKITTI}. 
This results in a total of 956 scenes of varying lengths, and overall approximately 864,000 frames.
Furthermore, to prevent overfitting to larger datasets, we apply uniform sampling across datasets. 
We simulate different frame rates by randomly sampling every first up to every fourth frame in a sequence.
Following VDA, we apply random cropping to a resolution of 512$\times$512. 

On top of VDA's training procedure, we utilise a simple frame augmentation technique to increase information exchange in the temporal dimension. For this, we select random rectangular-shaped regions in every frame and set the respective RGB values to 0. The number of affected pixels varies uniformly between 0\,\% and 40\,\% for every frame.  
Experiments show that this strategy improves accuracy (see \cref{subsec:SaSCon}).

\subsection{oVDA Loss Function}
\label{subsec:Loss}
VDA uses a linear combination of the scale- and shift-invariant loss (SSI) $\mathcal{L}_\textrm{SSI}^{\textrm{Image}}$, initially introduced in \mbox{MiDaS}~\cite{OriginalMidas}, and a newly proposed Temporal Gradient Matching loss $\mathcal{L}_\textrm{TGM}$~\cite{VDA}. Note that the SSI loss is modified by calculating the scale and shift parameters over sequences rather than over individual images, to encourage consistent depth within each sequence~\cite{VDA}, which we denote as $\mathcal{L}_\textrm{SSI}^{\textrm{Scene}}$. 
The total VDA loss is then defined as

\begin{equation}
	\mathcal{L}_\textrm{VDA} = \alpha \cdot \mathcal{L}_\textrm{SSI}^{\textrm{Scene}} + \beta \cdot \mathcal{L}_\textrm{TGM},
    \label{eq:LVDA}
\end{equation}
where $\alpha$ and $\beta$ are weighting parameters.

While this loss works well in the offline setting, it is suboptimal for our online setting. 
In VDA, the SSI loss is applied across the entire sequence, which aligns all frames jointly. 
As a result, each frame contributes equally to the scale, regardless of its temporal position in the sequence, pushing the model to predict scale-consistent batches. 
In an online setting, this is suboptimal since the model should predict the current frame at the same scale as preceding frames, resulting in a considerably more constrained task.

We address this issue by proposing the Scale-and-Shift Consistency Loss (SaSCon). 
First, we calculate the optimal scale and shift for the first frame in the sequence. 
This alignment (scale and shift) is then applied to the entire sequence. Simultaneously, we compute the optimal per-frame scale and shift to align each frame individually. 
Then, we calculate the $L_1$ loss between two depth maps: 

\begin{equation}
\mathcal{L}_\textrm{SaSCon} = \sum_{i=1}^{c} \left| \hat{D_i}^{\textrm{first}} - \hat{D_i}^{\textrm{indi.}} \right|_1,
\label{eq:SaSCon}
\end{equation}
where $\hat{D_i}^{\textrm{first}}$ is the predicted depth of the $i$-th frame aligned using the scale and shift derived from the first frame, and $\hat{D_i}^{\textrm{indi.}}$ is the predicted depth of the same frame aligned with its own per-frame scale and shift.
While the $\mathcal{L}_\textrm{SSI}^{\textrm{Scene}}$ learning objective aims to generate a globally consistent depth sequence, $\mathcal{L}_\textrm{SaSCon}$ focuses solely on temporal consistency, i.e. predicting a scale- and shift-consistent next frame. This loss is advantageous in our sliding-window-based online setting.
Our final training loss is defined as

\begin{equation}
\mathcal{L}_\textrm{oVDA} = \alpha \cdot \mathcal{L}_\textrm{SSI}^{\textrm{Scene}} + \beta \cdot \mathcal{L}_\textrm{TGM} + \gamma \cdot \mathcal{L}_\textrm{SaSCon},
\label{eq:LoVDA}
\end{equation}
where $\alpha = \beta = \gamma$ are chosen as 1.

\section{Experiments}
\label{sec:Experiments}
We evaluate our online method with respect to three different datasets in a zero-shot setting: KITTI~\cite{Kitti} (outdoor), Bonn~\cite{Bonn} (indoor), and Sintel~\cite{Sintel} (synthetic). In contrast to other online depth prediction approaches we evaluate results by computing the optimal scale and shift \emph{only} for the first frame and then align the entire depth video using these two parameters. Our evaluation is motivated by a typical, real-world use case where scale and shift is computed from external measurements or a calibration object. This evaluation is more challenging, as drifts in scale and/or shift can no longer be partially mitigated by global alignment. Also, in contrast to previous works such as VDA~\cite{VDA}, FlashDepth~\cite{flashdepth}, or DepthCrafter~\cite{DepthCrafter}, we evaluate video sequences in full length without limiting the number of frames, which is more realistic for an online setting.
Note, an evaluation with global alignment is in the supplement. 

\subsection{Zero-Shot Results}
\label{subsec:Zero-Shot Results}
After resizing to match the native resolution of each dataset, we compute the absolute relative error, defined as
$\text{AbsRel} = \frac{1}{N} \sum_i \frac{|D_i - \hat{D}_i^\prime|}{D_i}$
and the inlier ratio $\delta_1 = \frac{1}{N} \sum_i \mathbb{I} \left( \max \left( \frac{D_i}{\hat{D}_i^\prime}, \frac{\hat{D}_i^\prime}{D_i} \right) < 1.25 \right)$,
where $D_i$ is the ground truth depth, $\hat{D}_i^\prime$ the aligned predicted depth, $\mathbb{I}(\cdot)$ the indicator function, and $N$ the number of valid pixels.
Aligning only with respect to the first frame makes temporal consistency strongly correlated with the overall performance of each method, as a drifting scale and shift automatically lead to larger errors.
As in prior work \cite{VDA,flashdepth}, the following analysis focuses on the inlier $\delta_1$ ratio.
Results are shown in \cref{tab:ZeroShot}.
{
\setlength{\tabcolsep}{5.6pt}
\begin{table*}[t]
  \small
  \centering
  \begin{tabular}{lccccccccc|c}
    \toprule
     \multirow{2}{*}{Method}& \multicolumn{3}{c}{KITTI~\cite{Kitti} ($375\times 1242$)} 			& \multicolumn{3}{c}{Bonn~\cite{Bonn} ($480\times 640$)}   			& \multicolumn{3}{c}{Sintel~\cite{Sintel} ($436\times 1024$)}   & \multirow{2}{*}{$\delta_1$-Rank}\\
     					& Res. & AbsRel ($\downarrow$) & $\delta_1$ ($\uparrow$) 		& Res. & AbsRel ($\downarrow$) & $\delta_1$ ($\uparrow$) 	& Res. & AbsRel ($\downarrow$) & $\delta_1$ ($\uparrow$) & \\
     \midrule \midrule
    
    DAv2-s~\cite{depth_anything_v2}				& \scriptsize{378$\times$1246} & \textcolor{gray}{0.216} & \textcolor{gray}{0.655}	& \scriptsize{490$\times$644} & \textcolor{gray}{0.202} & \textcolor{gray}{0.712} & \scriptsize{378$\times$1246} & \textcolor{gray}{0.390} & \textcolor{gray}{0.514} & \textcolor{gray}{5.3} \\
    
    DAv2-l~\cite{depth_anything_v2}				& \scriptsize{378$\times$1246} & \textcolor{gray}{0.243} & \textcolor{gray}{0.610}	& \scriptsize{490$\times$644} & \textcolor{gray}{0.196} & \textcolor{gray}{0.721} & \scriptsize{378$\times$1246} & \textcolor{gray}{0.403} & \textcolor{gray}{0.521} & \textcolor{gray}{5.0} \\
    
    FlashDepth~\cite{flashdepth}		& \scriptsize{378$\times$1246} & 0.165 & 0.760 	& \scriptsize{490$\times$658} & 0.131 & 0.789 & \scriptsize{434$\times$1022} & \underline{0.370} & \textbf{0.564} & \underline{2.6} \\
    
    FlashDepth-s~\cite{flashdepth}	& \scriptsize{378$\times$1246} & \underline{0.156} & \underline{0.774} 	& \scriptsize{490$\times$658} & \textbf{0.116} & 0.848 & \scriptsize{434$\times$1022} & \textbf{0.355} & 0.532 & \underline{2.6} \\
    
    NVDS~\cite{NVDS}**				& \scriptsize{288$\times$896} & 0.490 & 0.380 		& \scriptsize{480$\times$896} & 0.259 & 0.602 				& \scriptsize{448$\times$896} & 0.440 & 0.413 & 6.6 \\
    
    CUT3R~\cite{CUT3R}			& \scriptsize{144$\times$512} & 0.238 & 0.690 			& \scriptsize{384$\times$512} & \textbf{0.116} & \underline{0.850}		& \scriptsize{208$\times$512} & 0.635 & 0.400 & 4.3 \\
    
    ChronoDepth~\cite{ChronoDepth}* 	& \scriptsize{256$\times$896} & 0.515 & 0.289 		& \scriptsize{480$\times$640} & 0.262 & 0.587 				& \scriptsize{448$\times$1024} & 0.563 & 0.355 & 8.0 \\
    \midrule	
    Our oVDA				& \scriptsize{280$\times$924} & \textbf{0.140} & \textbf{0.809} 	& \scriptsize{480$\times$640} & \underline{0.118} & \textbf{0.871} 	& \scriptsize{329$\times$924} & 0.380 & \underline{0.548} & \textbf{1.3} \\
    \bottomrule
  \end{tabular}
  \caption{Quantitative results for outdoor (KITTI~\cite{Kitti}), indoor (Bonn~\cite{Bonn}) and synthetic datasets (Sintel~\cite{Sintel}). The maximum depth is clipped to $80$m. We evaluate full sequences without a maximum frame cap. The average sequence length is $311$ for KITTI, $1118$ for Bonn, and 46 for Sintel. Predictions are resized to the original resolution before evaluation. 
  Depths are aligned only with respect to the first frame and the respective scale and shift are then applied to the entire video sequence. For each method and dataset we show the processing resolution alongside the Absolute Relative Error (AbsRel) and the inlier ratio ($\delta_1$). Depth Anything v2 small (DAv2-s) and DAv2-l are highlighted grey since they are single-frame methods and not designed to predict temporally consistent videos. *For online usage of ChronoDepth, we predict a single new frame with a context window of four. **We use the online version (no backwards refinement).}
  \label{tab:ZeroShot}
\end{table*}
}

We compare our approach against ChronoDepth~\cite{ChronoDepth}, a diffusion-based method, NVDS~\cite{NVDS}, a method which smooths single-frame depth predictions (details in the supplement),  FlashDepth~\cite{flashdepth} which augments a single-frame method with temporal blocks and CUT3R~\cite{CUT3R}, a persistent state model. 
Depth Anything v2~\cite{depth_anything_v2} is shown in grey, as it is a single-frame method with no temporal context but strong generalisation capabilities. It is still often applied to video depth prediction even though it can produce depths with inconsistent scale and shift. 

In the outdoor driving scenario (KITTI), our oVDA method outperforms the best competitor, FlashDepth-s, by more than 3\,pp in the $\delta_1$ inlier ratio, demonstrating that our sliding window approach is able to produce high accuracy, temporally consistent video depth predictions.
As well, in the indoor setting (Bonn), we outperform the best competitor, CUT3R, with a 2\,pp \(\delta_1\) gap.
Only for the synthetic Sintel datasets our method is the runner-up after FlashDepth. 
However, as we see next, FlashDepth needs more memory than our oVDA, since it has ten times more parameters, runs also slower and can still produce flickering depth (see \cref{fig:qualitative}).
Overall, the $\delta_1$-Rank clearly shows the superiority of our oVDA method. 

\cref{tab:Speed} presents the FPS and VRAM consumption of all evaluated methods. Since both metrics are highly dependent on the input resolution, we use the same resolution for all methods, even if they were not trained at that resolution. All measurements were done on an NVIDIA A100 GPU. 

FlashDepth-s achieves the highest frame rate, followed by our oVDA approach. Furthermore, oVDA achieves the lowest VRAM usage. Please note that, apart from the high frame rate, FlashDepth-s is inferior to oVDA in all other aspects: i) lower accuracy ($\delta_1$-Rank $1.3$ versus $2.6$), ii) higher VRAM usage, iii) temporally flickering depth (see \cref{fig:qualitative}),
and iv) lower temporal stability (see \cref{subsec:Temporal stability})

\subsection{Qualitative Comparison}
\label{subsec:Qualitative Comparison}
\cref{fig:qualitative} presents results for an in-the-wild video from DAVIS~\cite{DAVIS2017}.
We show four uniformly sampled frames, as well as a stitched image of 24-pixel-wide vertical slices (red column in each image). This helps to assess temporal consistency, where, ideally, the stitched image shows a smooth and stable depth/RGB gradient over time. 
Note that ChronoDepth~\cite{ChronoDepth} and NVDS~\cite{NVDS} are run in their online setting.
Our oVDA method achieves the highest temporal consistency, as seen in the stable depth gradient of the longboard rider (last row), followed by CUT3R~\cite{CUT3R}.
FlashDepth and FlashDepth-s~\cite{flashdepth} in particular produce noticeable flickering artifacts, making it less suitable for downstream applications \cite{OnlineDynamicDepth_2023}.
Additionally, our oVDA approach, as well as FlashDepth and FlashDepth-s, preserves fine details in the depth prediction, such as individual tree leaves. In contrast, CUT3R produces coarse and blurry depth maps.
These results highlight the ability of oVDA to deliver high-quality, temporally stable depth predictions suitable for demanding applications like augmented reality (AR).
For more visualisations we refer to the supplement and the supplementary videos.

{
\setlength{\tabcolsep}{2pt}
\begin{figure*}[h]
  \centering
  \begin{tabular}{@{}m{0.03\linewidth} m{0.97\linewidth}@{}}
    \centering\rotatebox{90}{RGB Video} &
    \includegraphics[width=\linewidth]{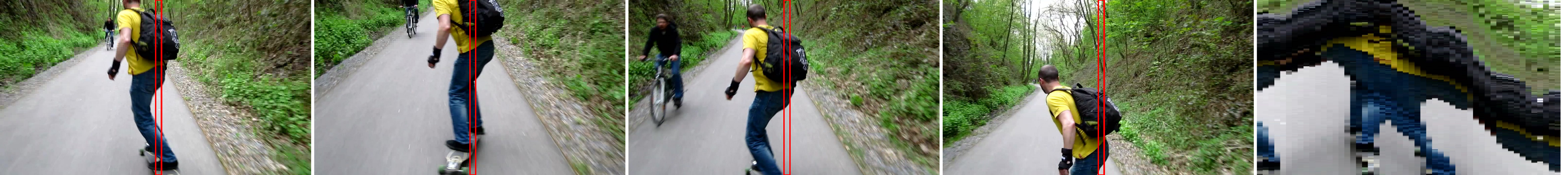} \\[-2pt]
    
    \centering\rotatebox{90}{NVDS} &
    \includegraphics[width=\linewidth]{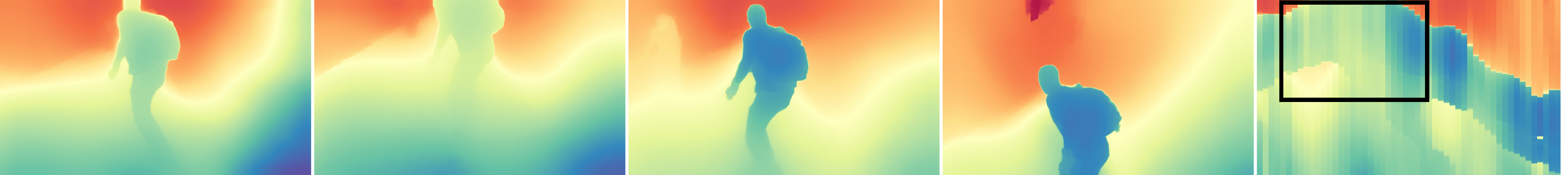} \\[-2pt]

    \centering\rotatebox{90}{ChronoDepth} &
    \includegraphics[width=\linewidth]{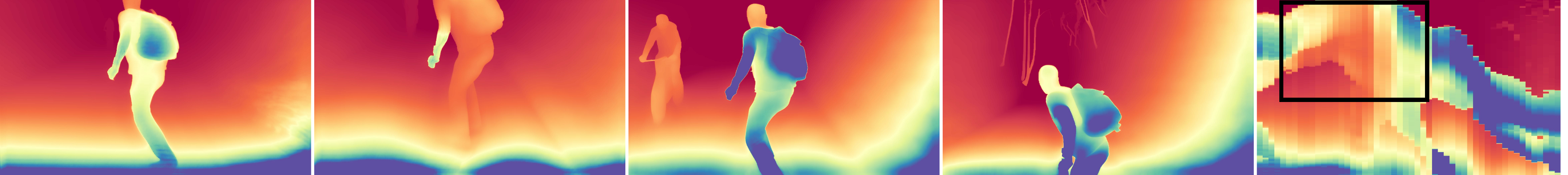} \\[-2pt]
    
    \centering\rotatebox{90}{CUT3R} &
    \includegraphics[width=\linewidth]{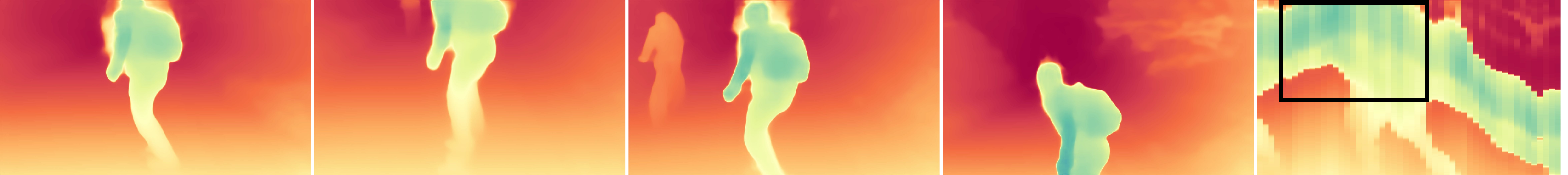} \\[-2pt]
    
    \centering\rotatebox{90}{FlashDepth} &
    \includegraphics[width=\linewidth]{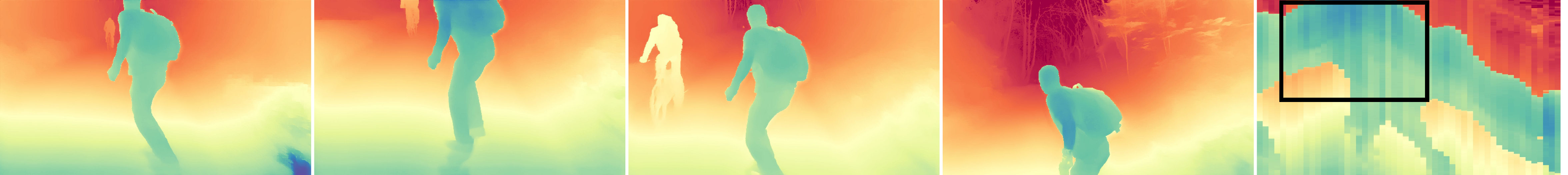} \\[-2pt]
  	
  	\centering\rotatebox{90}{FlashDepth-s} &
    \includegraphics[width=\linewidth]{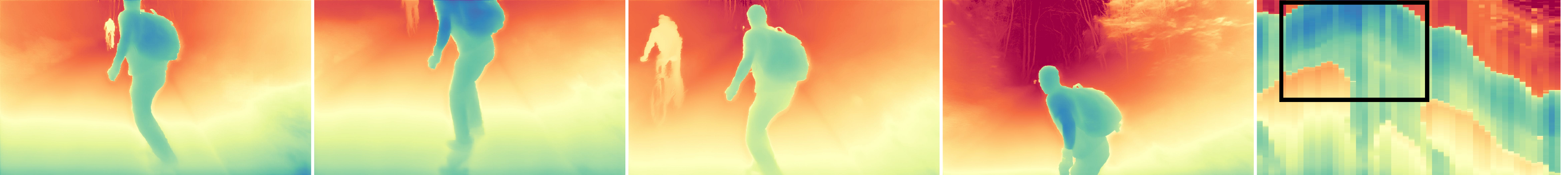} \\[-2pt]
    
    \centering\rotatebox{90}{Ours} &
    \includegraphics[width=\linewidth]{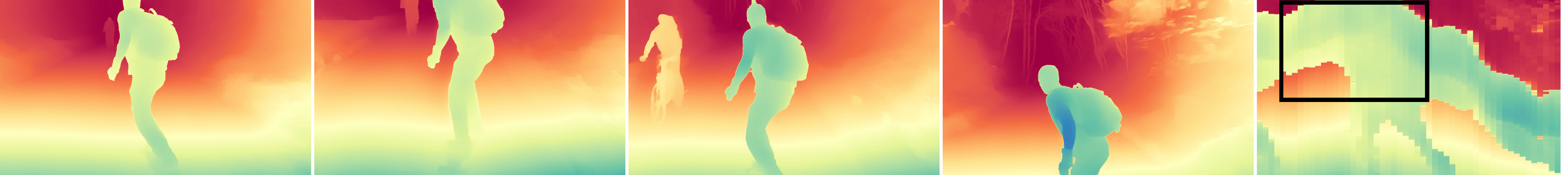} \\
  \end{tabular}
  \caption{Visual comparison for an in-the-wild-video from DAVIS~\cite{DAVIS2017}. We show four frames, equally spaced in time throughout the entire video sequence. The last row gives a stitched image of vertical slices (indicated by the red column in each RGB image, where the column is 24 pixels wide).The rightmost image is again the stitched version of the predicted depths to inspect temporal consistency. Note that both ChronoDepth~\cite{ChronoDepth} and NVDS~\cite{NVDS} are run in their online setting. We see that oVDA produce the temporally most consistent results followed by CUT3R~\cite{CUT3R}, in contrast to e.g., FlashDepth-s~\cite{flashdepth} (black box). However, the predictions of CUT3R are more blurry compared to ours. In summary, our approach is visually best in terms of temporal consistency and on par with FlashDepth in terms of details.}
  \label{fig:qualitative}
\end{figure*}
}

\begin{table}[h]
	\centering
    \small
	\begin{tabular}{l|lcc}
		Res. & Method & FPS ($\uparrow$) & VRAM [GB] ($\downarrow$)\\
		\midrule \midrule
		\multirow{8}{*}{\rotatebox{90}{280$\times$924}} 	& DAv2-s 		& \textcolor{gray}{72} & \textcolor{gray}{0.25} \\
		 				& DAv2-l 									& \textcolor{gray}{12} & \textcolor{gray}{2.04} \\
						& FlashDepth-s 								& \textbf{60} & \underline{0.69} \\
						& FlashDepth 								& 30 & 2.72 \\
						& NVDS* 										& 7 & 9.09 \\
						& CUT3R*										& 10 & 6.70 \\
						& ChronoDepth*								& 2 & 10.27 \\
						& Our oVDA										& \underline{42} & \textbf{0.45} \\	
	\end{tabular}
	\caption{Speed Comparison. All experiments were conducted on an NVIDIA A100 graphics card, processing 5,000 frames. We round the FPS to the closest integer and the VRAM to 10 MB. We try to keep the processing resolution as close as possible to 280$\times$924. * marks methods that do not support this resolution and use a slightly adjusted resolution.}
	\label{tab:Speed}
\end{table}

\subsection{Edge Device}
\label{subsec:Edge Device}
{
\setlength{\tabcolsep}{5pt}
\begin{table}[h]
	\centering
    \small
	\begin{tabular}{lcccc}
		Precision & AbsRel ($\downarrow$) & $\delta_1$ ($\uparrow$) & FPS ($\uparrow$) & VRAM [GB] ($\downarrow$) \\
		\midrule \midrule
		FP32  & 0.1399 & 0.8094 & 9 & 0.64\\
		FP16  & 0.1398 & 0.8093 & 20 & 0.49\\
	\end{tabular}
	\caption{Results of oVDA on an NVIDIA Jetson Orin NX. We achieve 20\,FPS with 16-bit floating-point (FP16) precision. We evaluated two precisions for KITTI~\cite{Kitti} and observe that FP16 does not significantly degrade performance. FPS and VRAM usage were measured at a resolution of 280$\times$924 over 1,000 frames.}
	\label{tab:EdgeDevice}
\end{table}
}
To demonstrate oVDA’s versatility and suitability for edge device deployment, we run it on an NVIDIA Jetson Orin NX, which is a compact device with limited VRAM and limited GPU power. As shown in \cref{tab:EdgeDevice}, oVDA reaches 9\,FPS with FP32, sufficient for tasks such as robotic indoor navigation or obstacle avoidance. When using FP16, the runtime increases to 20\,FPS without a major drop in accuracy, enabling more demanding applications like real-time AR or low-latency drone control. This stability across precisions likely stems from the fact that a frozen encoder is used and no numerically extreme activations occur in the Motion Module, making oVDA ideal for resource-constrained hardware.

It is worth noting that oVDA outperforms methods like CUT3R, which has 26$\times$ more parameters, in terms of accuracy (see \cref{tab:ZeroShot}), runtime, and VRAM (see \cref{tab:Speed}). In particular, oVDA achieves twice the FPS (20 versus 10) on an NVIDIA Jetson Orin NX compared to CUT3R on an NVIDIA A100 graphics card.   

We will release model weights, code, and a compilation script for edge device deployment and integration into downstream applications.
 
\section{Ablation Studies}
\label{sec:AblationStudies}
We demonstrate in \cref{subsec:SaSCon} the benefits of our proposed SaSCon loss. \cref{subsec:Comparison to VDA} compares our oVDA method to the \mbox{offline} base model Video Depth Anything (VDA)~\cite{VDA}. Finally,  an ablation study with respect to temporal stability is given in 
\cref{subsec:Temporal stability}.
Note, an ablation of different context lengths is available in the supplement.
In the following, all results are based on the KITTI dataset~\cite{Kitti}. As in \cref{sec:Experiments}, the alignment of the predicted depth sequence is performed with respect to the first frame.

\subsection{SaSCon Loss}
\label{subsec:SaSCon}
\cref{tab:Losses} presents results for various loss configurations. 
Applying the oVDA inference scheme to VDA without fine-tuning produces poor performance, as VDA is not designed for online inference and its attention mechanisms assume that past frames can be modified. Fine-tuning with the original $\mathcal{L}_{\textrm{VDA}}$ loss (see \cref{eq:LVDA}) improves results by nearly 10\,pp over the untrained baseline. The frame augmentation strategy described in \cref{subsec:Training} yields a further gain of about 1\,pp. 
Adding our proposed $\mathcal{L}_\textrm{SaSCon}$ loss (see \cref{eq:SaSCon}) on top of $\mathcal{L}_{\textrm{VDA}}$ yields an additional improvement of approximately 1.7\,pp. Finally, our $\mathcal{L}_{\textrm{oVDA}}$ loss (see \cref{eq:LoVDA}) combined with frame augmentation delivers the highest overall performance, defining our proposed oVDA method.
\begin{table}[h]
	\centering
	\begin{tabular}{lcc}
		Loss &  AbsRel ($\downarrow$) & $\delta_1$ ($\uparrow$) \\
		\midrule \midrule
		No Training 				& 0.201 & 0.693 \\
		$\mathcal{L}_\textrm{VDA}$ 	& 0.155	& 0.778 \\
		$\mathcal{L}_\textrm{VDA}$ + Frame Aug. & 0.149 & 0.788 \\
		$\mathcal{L}_\textrm{VDA}$ + $\mathcal{L}_\textrm{SaSCon}$ & 0.148 & 0.795 \\	
	  $\mathcal{L}_\textrm{oVDA}$ + Frame Aug. & 0.140 & 0.809 \\	
	\end{tabular}
	\caption{Accuracy with respect to varying loss configurations. We observe that our proposed $\mathcal{L}_\textrm{SaSCon}$ loss improves accuracy. Our final oVDA method with $\mathcal{L}_{\textrm{oVDA}}$ loss in conjunction with frame augmentation yields best performance.}
	\label{tab:Losses}
\end{table}
{
\setlength{\tabcolsep}{5pt}
\begin{table}[b]
	\centering
	\begin{tabular}{lccc}
		Method & AbsRel ($\downarrow$) & $\delta_1$ ($\uparrow$)& Latency [ms] ($\downarrow$) \\
		\midrule \midrule
		VDA-l~\cite{VDA} 	& 0.127 		& 0.860 		& 579 \\
		VDA-s~\cite{VDA} 	& 0.137 		& 0.832 		& 232 \\
		Ours ($c=32$)		& 0.136 		& 0.811 		& 26 \\	
	\end{tabular}
	\caption{Comparison to Video Depth Anything~\cite{VDA} at a resolution of 280$\times$924. In contrast to VDA, our oVDA method has no access to future frames which reduces accuracy slightly.
    Since VDA predicts in batches, we provide the batch-latency of VDA. The latency of online oVDA is about $9\times$ less than that of offline VDA-s.} 
	\label{tab: VDA-Comparison}
\end{table}
}
\subsection{Comparison to VDA}
\label{subsec:Comparison to VDA}
The offline Video Depth Anything (VDA)~\cite{VDA} approach defines 
an upper bound for our method, illustrated in \cref{tab: VDA-Comparison}
Since VDA uses a fixed batch size (i.e. context window) of 32, we compare it to our $c = 32$ model. VDA exists in two variants, i.e. -small and -large, where the latter has about $13\times$ more parameters. Note that our oVDA model builds upon VDA-s. oVDA lags behind VDA-s by only 2.2\,pp in accuracy ($\delta_1$), which can be expected since it operates without access to future frames. The difference between an offline and an online version can be seen in the latency. The latency of VDA-s is about $9 \times$ higher than that of oVDA.

\subsection{Temporal Stability}
\label{subsec:Temporal stability}
\begin{figure}
	\centering
   	\includegraphics[width=\linewidth]{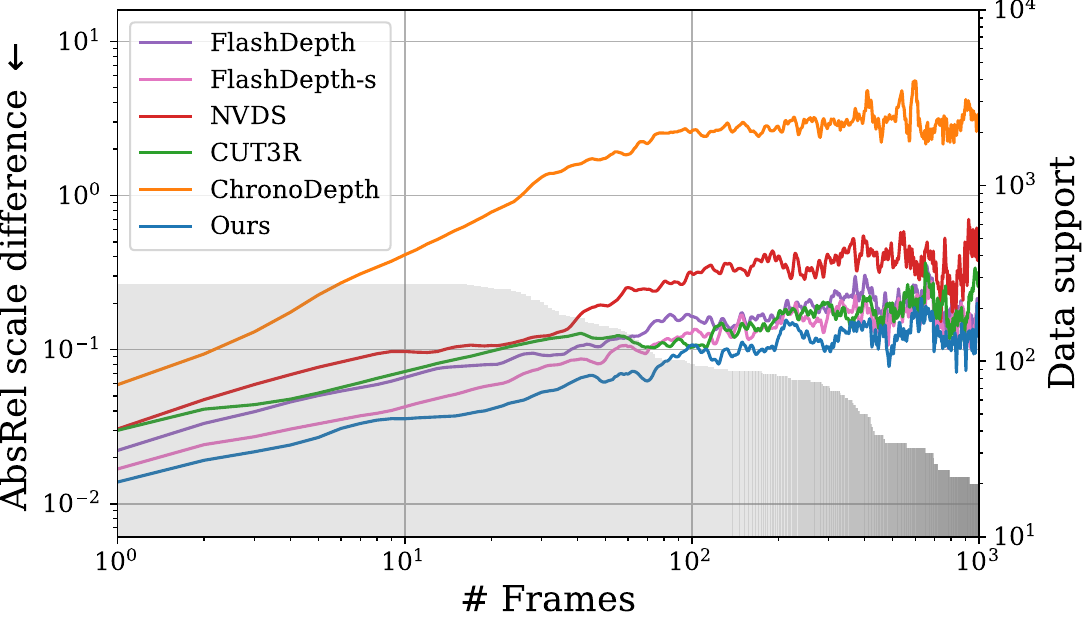}

   \caption{Scale drift over time for KITTI~\cite{Kitti}. We plot the absolute relative difference between the optimal scale for the first frame and the optimal scale for each subsequent frame. The grey histogram gives the data support, i.e. the number of data points used to calculate the scale error, which decreases for later frames due to varying sequence lengths. We smoothed the scale error with a window size of 4 for better visualisation. Our oVDA method has lowest scale drift.}
   \label{fig:Scale Drift}
\end{figure}

Maintaining a temporal consistent scale is essential for online video depth estimation, as alignment typically requires either costly external measurements or heavy computations~\cite{PointBasedFusion,VDA,align3r}.
\cref{fig:Scale Drift} visualises the scale drift over time by plotting the absolute relative difference between the optimal scale $s_0$ for the first frame 0 and the optimal scale $s_i$ computed for each subsequent frame $i$: AbsRel = 
$\frac{1}{N_j} \sum_j \frac{|s_0 - s_j|}{s_0}$. We refer to $N_j$ as data support (grey area in \cref{fig:Scale Drift}), defined as the number of sequences that contain at least $j$ frames. 
We observe that our oVDA method consistently exhibits lower scale drift than all competitors, particularly for up to 300 frames. Beyond this point, FlashDepth and CUT3R approach the performance of oVDA, while NVDS and ChronoDepth continue to show a substantially higher drift. After 1000 frames, data support drops below 20 sequences, making further analysis less reliable.

\section{Conclusion and Limitations}
\label{sec:Conclusion}
We presented oVDA, a transformation of the offline VDA technique to the online setting, inspired by inference and training strategies used in LLMs. Extensive experiments demonstrate that oVDA achieves state-of-the-art accuracy and temporal consistency across both indoor and outdoor datasets, while significantly reducing memory usage. Furthermore, it achieves 42 FPS on an NVIDIA A100 and 20 FPS on an NVIDIA Jetson edge device.
Nevertheless, oVDA still struggles with scale drift for extremely long video sequences. Moreover, fast-moving objects can produce so-called trailing artefacts, highlighting the need for further research, especially for safety-critical applications.

\section{Acknowledgements}
\label{sec:Acknowledgements}
This project has been supported by HARMAN International.

{
    \small
    \bibliographystyle{ieeenat_fullname}
    \bibliography{main}
}
\appendix
\clearpage
\setcounter{page}{1}
\maketitlesupplementary

\section{Supplementary Material}
\label{sec:Supplementary}
Unless otherwise stated, all results presented in this section are obtained using the complete KITTI dataset~\cite{Kitti}, where the alignment was performed with respect to the first frame.

\subsection{Global Alignment}
\label{subsec:globalalignemnt}
So far all evaluations were done by computing the scale and shift with respect to the first frame of a sequence and then use these parameters to adjust the full sequence. In \cref{tab:ZeroShotGlobalAlign} we give results with respect to {\bf global} alignment. 
This means that we evaluate and calculate the optimal scale and shift either on the first 500 frames or on all frames. 
Aligning globally partially compensates for scale drift, resulting in overall improved metrics for all methods.

\subsection{Context Length}
\label{subsec:Context length}
\cref{tab:Context Lenght} reports the performance of oVDA for different context lengths $c$. As expected, shorter context windows yield faster inference at the cost of reduced accuracy. Throughout our experiments, we adopt $c=16$ as the default setting, as it provides a favourable trade-off between efficiency and accuracy.

We also explored simulating longer context windows by maintaining multiple caches of frames. To remain consistent with training, each cache stores temporally equally spaced frames. By iteratively rotating through these caches, we can effectively double, triple, or quadruple the context length, exposing the network to every second, third, or fourth frame, respectively. To maximise temporal consistency, we first predict frames sequentially until the maximum context length $c$ is reached, after which the cached frames are distributed across the additional caches and updated in an alternating fashion. In principle, this strategy should allow for a flexible balance between memory usage and accuracy, while keeping the FPS unchanged.

However, as shown in \cref{tab:VirtualContext}, this approach does not yield performance improvements in practice. We hypothesise that the use of separate caches introduces independent scale and shift drifts across caches, which ultimately degrades overall accuracy.
{
\setlength{\tabcolsep}{5pt}
\begin{table}[h]
	\centering
  \small
	\begin{tabular}{lcccc}
		Context & AbsRel ($\downarrow$) & $\delta_1$ ($\uparrow$) & FPS ($\uparrow$) & VRAM [GB] ($\downarrow$) \\
		\midrule \midrule
		8 & 0.146 & 0.800 & 43 & 0.37\\
		16 (Ours) & 0.140 & 0.809 & 42 & 0.45 \\
		32 & 0.136 & 0.811 & 39 & 1.36 \\		
	\end{tabular}
	\caption{Accuracy and speed across different context lengths. A length of 32 provides only a marginal accuracy gain at the cost of reduced frame rate, while a length of 8 does not yield a meaningful improvement in runtime. We therefore adopt 16 as a balanced compromise.}
	\label{tab:Context Lenght}
\end{table}
}
{
\setlength{\tabcolsep}{5pt}
\begin{table}[h]
	\centering
  \small
	\begin{tabular}{lcccc}
		Context & AbsRel ($\downarrow$) & $\delta_1$ ($\uparrow$) & FPS ($\uparrow$) & VRAM [GB] ($\downarrow$) \\
		\midrule \midrule
        8             & \textbf{0.146} & \textbf{0.800} & 43 & 0.37\\
        Virtual 16    & 0.149 & 0.793 & 43 & 0.45 \\
        \hline
		16            & \textbf{0.140} & \textbf{0.809} & 42 & 0.45 \\
		Virtual 32    & 0.141 & 0.806 & 42 & 0.58 \\
		Virtual 48    & 0.142 & 0.802 & 42 & 0.68 \\	
        Virtual 64    & 0.143 & 0.800 & 42 & 0.72 \\
	\end{tabular}
	\caption{Performance of oVDA when virtually extending the context length using multiple caches. Starting from oVDA with a context of 8, we double the context (Virtual 16) but do not reach the accuracy of the full oVDA with $c=16$. Repeating the experiment with the $c=16$ model, we double, triple, and quadruple the context length via additional caches, yet still observe a performance drop. We attribute this to scale and shift drifting independently across caches.}
	\label{tab:VirtualContext}
\end{table}
}

\begin{table*}[t]
  \scriptsize
  \centering
  \begin{tabular}{lccccccccc|c}
    \toprule
     \multirow{3}{*}{Method}& \multicolumn{3}{c}{KITTI~\cite{Kitti} ($375\times 1242$)} 			& \multicolumn{3}{c}{Bonn~\cite{Bonn} ($480\times 640$)}   			& \multicolumn{3}{c}{Sintel~\cite{Sintel} ($436\times 1024$)}   & \multirow{2}{*}{$\delta_1$-Rank}\\
     					& Res. & AbsRel ($\downarrow$) & $\delta_1$ ($\uparrow$) 		& Res. & AbsRel ($\downarrow$) & $\delta_1$ ($\uparrow$) 	& Res. & AbsRel ($\downarrow$) & $\delta_1$ ($\uparrow$) & \\
            & & 500 / all & 500 / all & & 500 / all & 500 / all & & all & all & 500 / all \\
     \midrule \midrule
    
    DAv2-s~\cite{depth_anything_v2}	& \scriptsize{378$\times$1246} & \textcolor{gray}{0.128 / 0.146} & \textcolor{gray}{0.836 / 0.786}	& \scriptsize{490$\times$644} & \textcolor{gray}{0.118 / 0.154} & \textcolor{gray}{0.867 / 0.776} & \scriptsize{378$\times$1246} & \textcolor{gray}{0.315} & \textcolor{gray}{0.579} & \textcolor{gray}{5.0 / 5.3} \\
    
    DAv2-l~\cite{depth_anything_v2}	& \scriptsize{378$\times$1246} & \textcolor{gray}{0.132 / 0.156} & \textcolor{gray}{0.816 / 0.753}	& \scriptsize{490$\times$644} & \textcolor{gray}{0.118 / 0.152} & \textcolor{gray}{0.869 / 0.785} & \scriptsize{378$\times$1246} & \textcolor{gray}{0.293} & \textcolor{gray}{0.598} & \textcolor{gray}{4.3 / 4.6} \\
    
    FlashDepth~\cite{flashdepth}		& \scriptsize{378$\times$1246} & 0.125 / 0.136 & \underline{0.856} / 0.825 	& \scriptsize{490$\times$658} & 0.084 / 0.119 & 0.944 / 0.856 & \scriptsize{434$\times$1022} & \underline{0.292} & \textbf{0.618} & \underline{2.0} / \underline{2.6} \\
    
    FlashDepth-s~\cite{flashdepth}	& \scriptsize{378$\times$1246} & \underline{0.121} / \underline{0.132} & 0.854 / \underline{0.828} 	& \scriptsize{490$\times$658} & \underline{0.075} / 0.110 & \textbf{0.961} / 0.876 & \scriptsize{434$\times$1022} & \textbf{0.282} & 0.586 & 2.6 / 3.0 \\

    NVDS~\cite{NVDS}**				& \scriptsize{288$\times$896} & 0.222 / 0.243 & 0.634 / 0.587 		& \scriptsize{480$\times$896} & 0.189 / 0.214 & 0.710 / 0.638 & \scriptsize{448$\times$896} & 0.364 & 0.489 & 7.3 / 7.6 \\
    
    NVDS-DPT~\cite{NVDS}**	& \scriptsize{288$\times$896} & 0.224 / 0.246 & 0.632 / 0.579 & \scriptsize{480$\times$896} & 0.181 / 0.206 & 0.725 / 0.652 & \scriptsize{448$\times$896} & 0.354 & 0.518 & 7.0 / 7.3 \\

    CUT3R~\cite{CUT3R}			& \scriptsize{144$\times$512} & 0.126 / 0.134 & 0.841 / 0.817 			& \scriptsize{384$\times$512} & \textbf{0.071} / \textbf{0.094} & \underline{0.960} / \textbf{0.908}		& \scriptsize{208$\times$512} & 0.322 & 0.536 & 4.0 / 3.6 \\
    
    ChronoDepth~\cite{ChronoDepth}* 	& \scriptsize{256$\times$896} & 0.251 / 0.297 & 0.548 / 0.471	& \scriptsize{480$\times$640} & 0.259 / 0.266 & 0.510 / 0.503 				& \scriptsize{448$\times$1024} & 0.531 & 0.308 & 8.6 / 9.0 \\
    \midrule	
    Our oVDA				& \scriptsize{280$\times$924} & \textbf{0.103} / \textbf{0.112} & \textbf{0.894} / \textbf{0.876} 	& \scriptsize{480$\times$640} & 0.079 / \underline{0.109} & \textbf{0.961} / \underline{0.887} 	& \scriptsize{329$\times$924} & 0.294 & \underline{0.604} & \textbf{1.3} / \textbf{1.6} \\
    \bottomrule
  \end{tabular}
  \caption{Quantitative results on outdoor (KITTI~\cite{Kitti}), indoor (Bonn~\cite{Bonn}), and synthetic (Sintel~\cite{Sintel}) datasets. The maximum depth is clipped to 80\,m. For this experiment, evaluation and {\bf alignment are performed globally on either the first 500 frames or on all frames}. Aligning globally on all evaluated frames partially compensates for scale drift, resulting in overall improved metrics. The $\delta_1$ ranking and general trends remain consistent with \cref{tab:ZeroShot}, although the larger FlashDepth~\cite{flashdepth} now slightly outperforms FlashDepth-s. *To adapt ChronoDepth to the online setting, we predict a single new frame with a context window of four. **For this method, we use the online implementation (without backward refinement).}
  \label{tab:ZeroShotGlobalAlign}
\end{table*}

\subsection{NVDS Details}
\label{sec:NVDS-Details}
The NVDS framework~\cite{NVDS} exists in two variants, differing in the choice of the single-image depth estimation backbone: MiDaS~\cite{Midas3.1} or DPT~\cite{DPT}.
Throughout our work, we primarily report results for NVDS-MiDaS, as this variant achieves the best performance on KITTI~\cite{Kitti} while also being computationally more efficient (MiDaS is approximately three times smaller than DPT).

Although NVDS-DPT yields slightly higher accuracy on other benchmarks, improving \(\delta_1\) by 2\,pp on Bonn and by 3.3\,pp on Sintel, these differences do not alter the overall ranking reported in \cref{tab:ZeroShot}. For completeness, we also include NVDS-DPT results under the global alignment setting in \cref{tab:ZeroShotGlobalAlign}.

\subsection{Comparison to VDA}
\label{subsec:ComparisontoVDA_supp}
In \cref{tab: VDA-Comparison-sup}, we report the FPS and VRAM usage of VDA compared to oVDA. As discussed in \cref{subsec:Comparison to VDA}, VDA exhibits significantly lower latency because it processes frames in batches. This batch-wise design also explains why the raw FPS of VDA is more than twice that of oVDA, as it benefits from strong parallelisation.

This parallelism not only accounts for VDA-s running at more than double the speed of our online variant, but also clarifies why Depth Anything v2 small (DAv2-s) achieves 72\,FPS, while VDA-s reaches 95\,FPS despite having a larger parameter count. To enable a fairer comparison, we adapted VDA to process frames individually. In this setting, the parallelization advantage disappears, and the frame rate drops to 43\,FPS; comparable to oVDA and even slower than DAv2-s.

The remaining gap of roughly 4\,FPS arises from redundant computations in our method. For each new frame, the keys and values of all context frames must be recomputed. This step is unavoidable, since latent features need to be cached before the $K/V$ linear layers, as positional encodings are applied before them and must be reapplied for every new frame.
{
\setlength{\tabcolsep}{4.8pt}
\begin{table}[h]
	\centering
	\begin{tabular}{lcccc}
		Method & $\delta_1$ ($\uparrow$)& FPS ($\uparrow$) & VRAM [GB] ($\downarrow$)\\
		\midrule \midrule
		VDA-l~\cite{VDA} 		& 0.860 		& 38 & 7.67 \\
		VDA-s~\cite{VDA} 		& 0.832 		& 95 & 1.76 \\
		Ours ($c=32$)			& 0.811 		& 39 & 1.36 \\	
	\end{tabular}
	\caption{Comparison with Video Depth Anything~\cite{VDA} at a resolution of 280$\times$924. VDA processes frames in batches, which allows for more efficient context handling and improved predictions by leveraging future frames, both of which are not accessible in an online setting. The FPS gap between our oVDA and VDA-s arises from the computational advantages of batch processing, as well as the need for oVDA to recompute $K/V$ for all context frames due to the positional encoding.}
	\label{tab: VDA-Comparison-sup}
\end{table}
}

\subsection{Motion Module}
\begin{figure*}[ht]
  \centering
   \includegraphics[width=\linewidth]{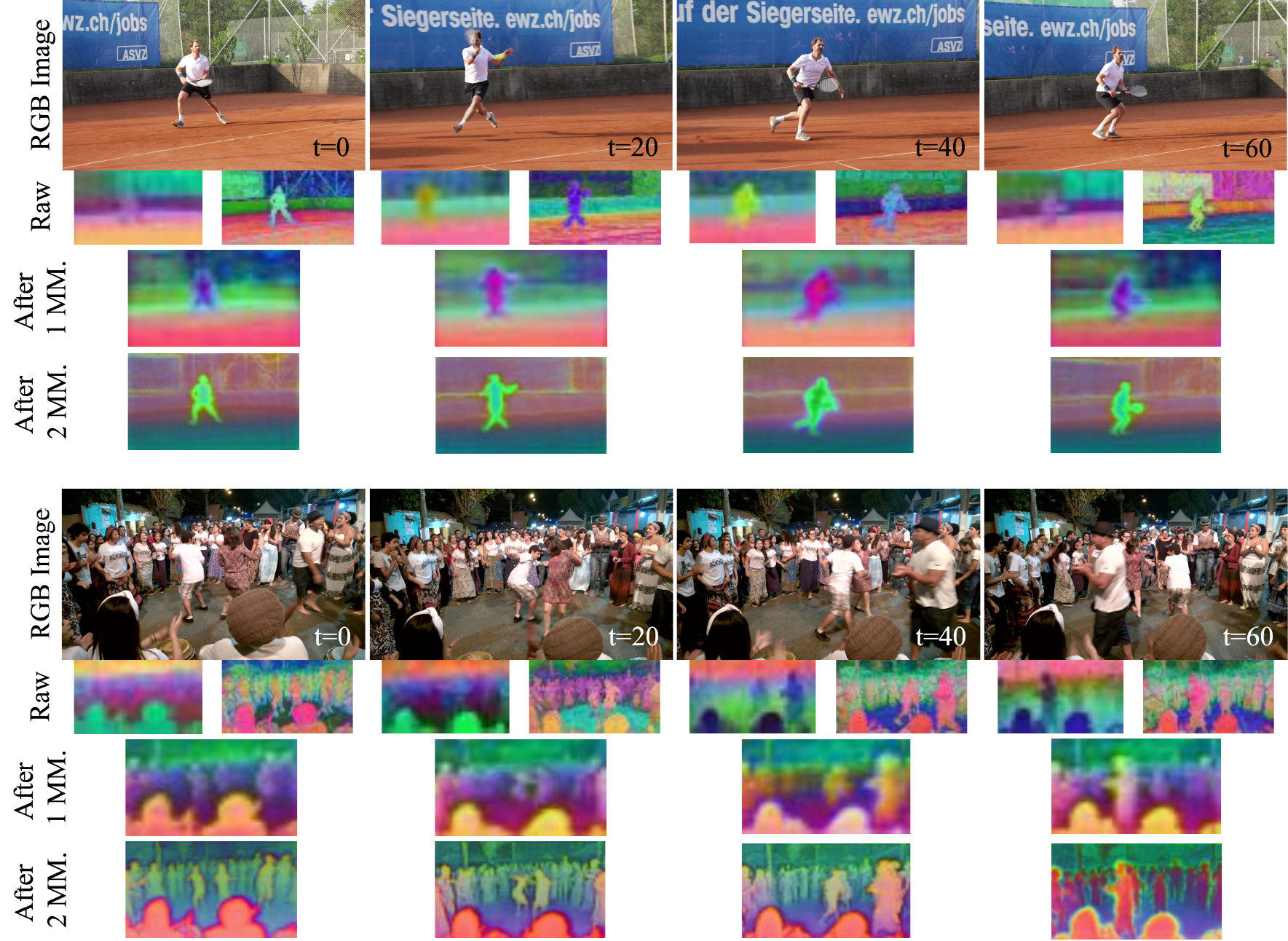}

   \caption{Visualisation of latent features using PCA. Raw encoder features ($F_1$, $F_2$) are temporally inconsistent, while features after one and two Motion Modules (MM.) remain stable across time, indicating that the Motion Modules enforce temporal consistency.}
   \label{fig:StableLatents}
\end{figure*}

To better understand the contribution of the Motion Modules (MMs) within our network, we visualise their effect in \cref{fig:StableLatents}. To render the latent features \(L_j\), we follow the procedure introduced in DINOv2: principal component analysis (PCA) is applied along the channel dimension, and the first three principal components are mapped to the RGB channels of the visualisation, while the remaining components are discarded.

In \cref{fig:StableLatents}, we present four temporally spaced RGB frames alongside the raw latent features extracted from the DINOv2 encoder, ($F_1$) and ($F_2$). Below these, we show the inputs to the final two Motion Modules (see \cref{fig:Architecture} in the main article for reference). These correspond to latent features ($L_j$) after processing by a single Motion Module (“After 1 MM”) and after two Motion Modules (“After 2 MMs”), respectively.

The visualisations clearly demonstrate that temporal consistency improves substantially once the features are processed by the Motion Modules. In particular, the colours of the ground and human figures remain far more stable across time compared to the raw encoder features. This confirms that the Motion Modules fulfil their intended role: enforcing temporal coherence in the latent representations.

\subsection{Further Visual Results}
To further illustrate the temporal consistency and overall quality of our approach, we provide additional results on in-the-wild DAVIS~\cite{DAVIS2017} videos in \cref{fig:sub_qualitative_coala,fig:qualitative_tennis,fig:qualitative_hike,fig:qualitative_duck}. The corresponding video files are included in the supplementary material as MP4s.

{
\setlength{\tabcolsep}{2pt}
\begin{figure*}[h]
  \centering
  \begin{tabular}{@{}m{0.03\linewidth} m{0.97\linewidth}@{}}
    \centering\rotatebox{90}{RGB Video} &
    \includegraphics[width=\linewidth]{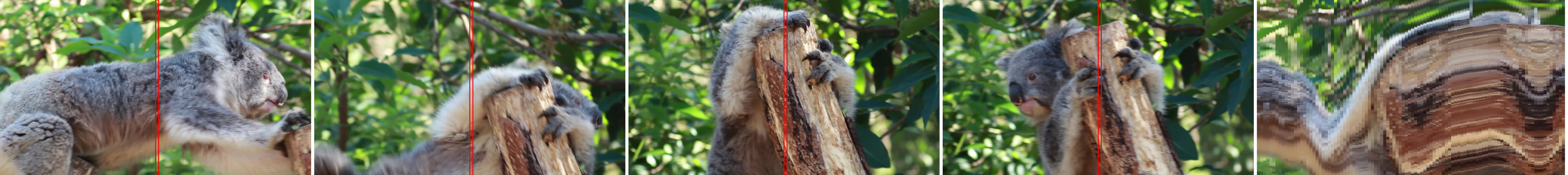} \\[-2pt]
    
    \centering\rotatebox{90}{NVDS} &
    \includegraphics[width=\linewidth]{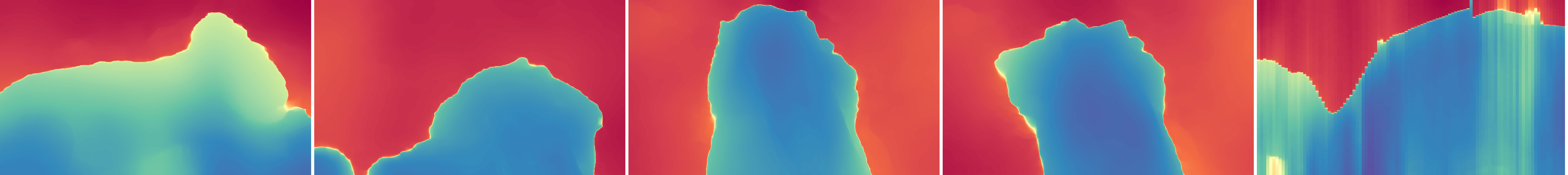} \\[-2pt]

    \centering\rotatebox{90}{ChronoDepth} &
    \includegraphics[width=\linewidth]{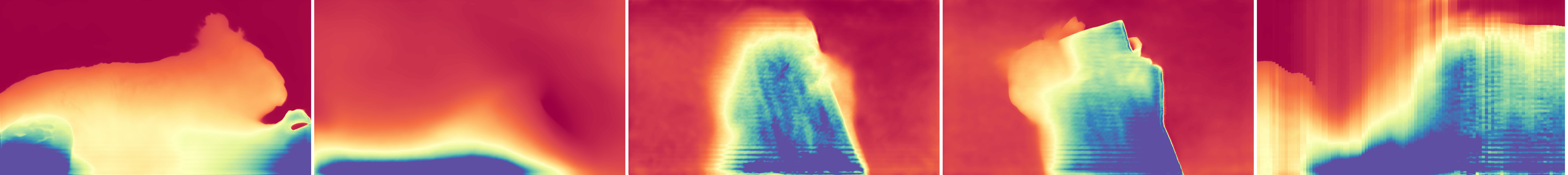} \\[-2pt]
    
    \centering\rotatebox{90}{CUT3R} &
    \includegraphics[width=\linewidth]{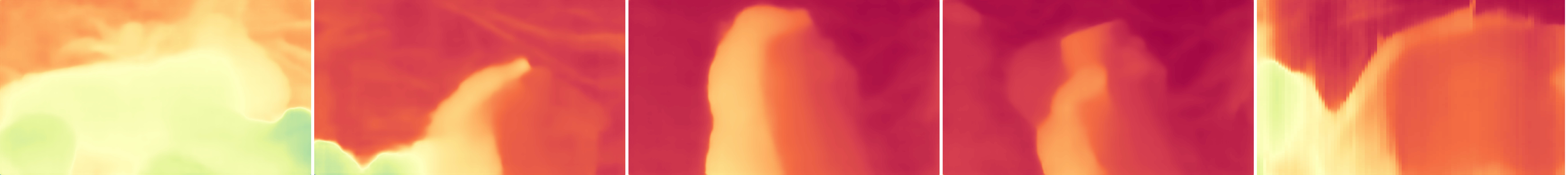} \\[-2pt]
    
    \centering\rotatebox{90}{FlashDepth} &
    \includegraphics[width=\linewidth]{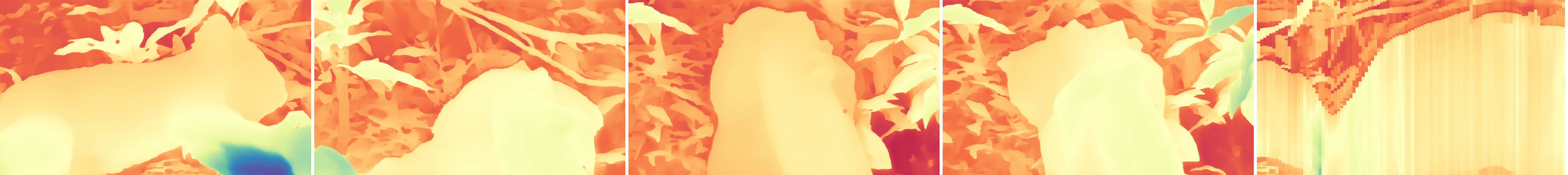} \\[-2pt]
  	
  	\centering\rotatebox{90}{FlashDepth-s} &
    \includegraphics[width=\linewidth]{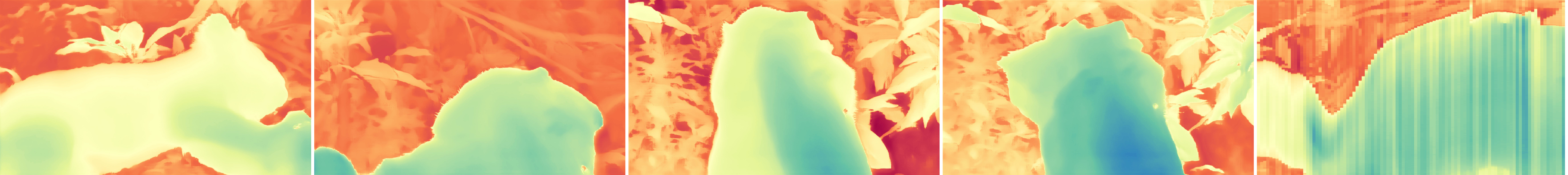} \\[-2pt]
    
    \centering\rotatebox{90}{Ours} &
    \includegraphics[width=\linewidth]{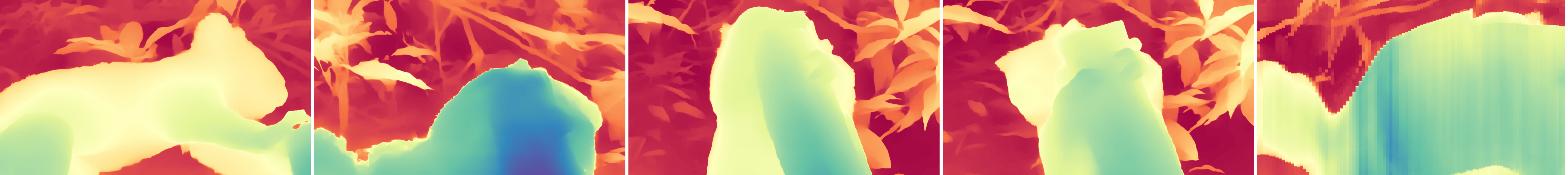} \\
  \end{tabular}
  \caption{NVDS~\cite{NVDS}, ChronoDepth~\cite{ChronoDepth}, and CUT3R~\cite{CUT3R} produce only a coarse estimate of the scene. FlashDepth~\cite{flashdepth}, FlashDepth-s~\cite{flashdepth} and our oVDA method yield more accurate predictions, although FlashDepth and FlashDepth-s still show noticeable flickering, particularly along the tree trunk.}
  \label{fig:sub_qualitative_coala}
\end{figure*}
}
{
\setlength{\tabcolsep}{2pt}
\begin{figure*}[h]
  \centering
  \begin{tabular}{@{}m{0.03\linewidth} m{0.97\linewidth}@{}}
    \centering\rotatebox{90}{RGB Video} &
    \includegraphics[width=\linewidth]{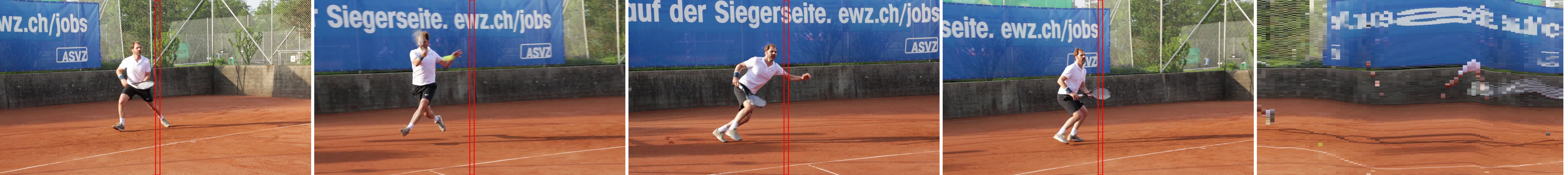} \\[-2pt]
    
    \centering\rotatebox{90}{NVDS} &
    \includegraphics[width=\linewidth]{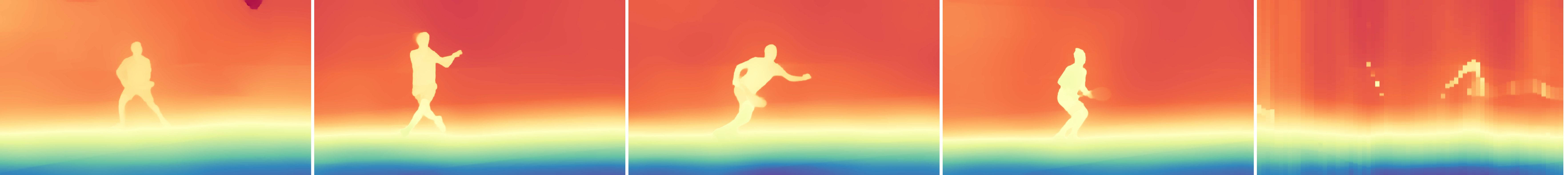} \\[-2pt]

    \centering\rotatebox{90}{ChronoDepth} &
    \includegraphics[width=\linewidth]{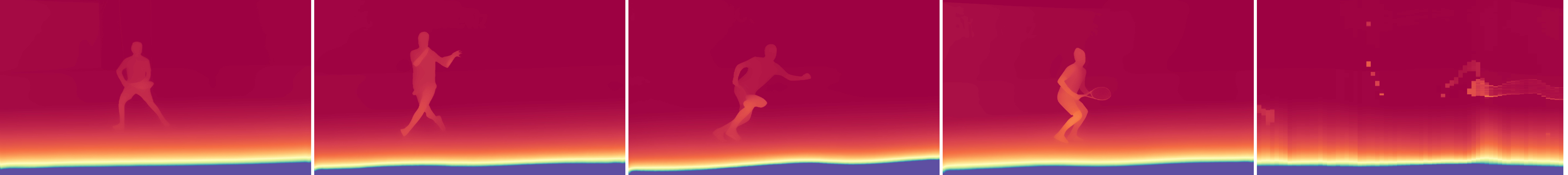} \\[-2pt]
    
    \centering\rotatebox{90}{CUT3R} &
    \includegraphics[width=\linewidth]{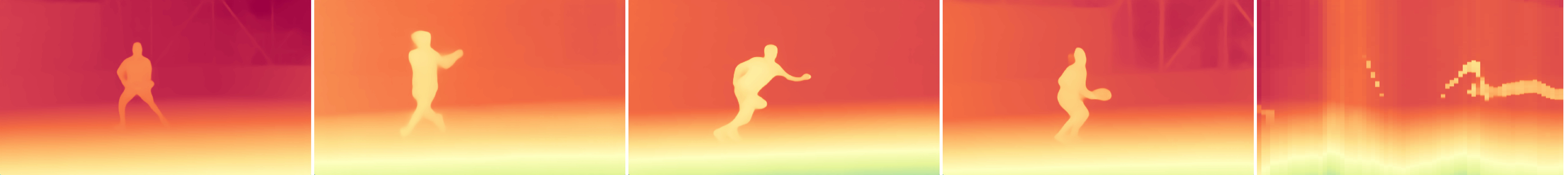} \\[-2pt]
    
    \centering\rotatebox{90}{FlashDepth} &
    \includegraphics[width=\linewidth]{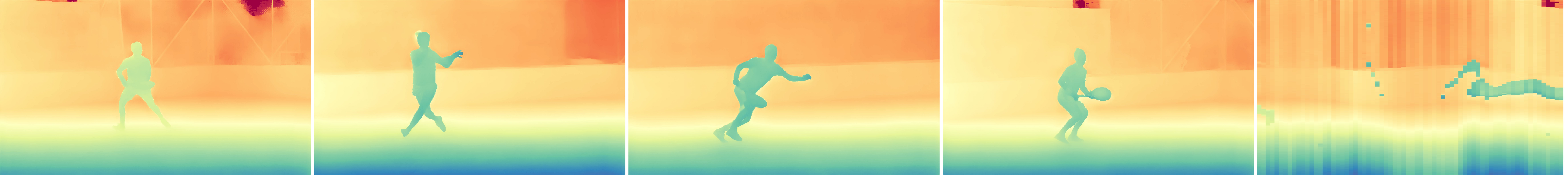} \\[-2pt]
  	
  	\centering\rotatebox{90}{FlashDepth-s} &
    \includegraphics[width=\linewidth]{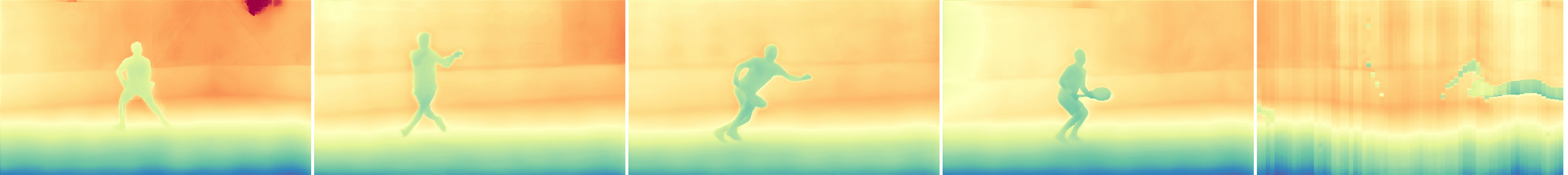} \\[-2pt]
    
    \centering\rotatebox{90}{Ours} &
    \includegraphics[width=\linewidth]{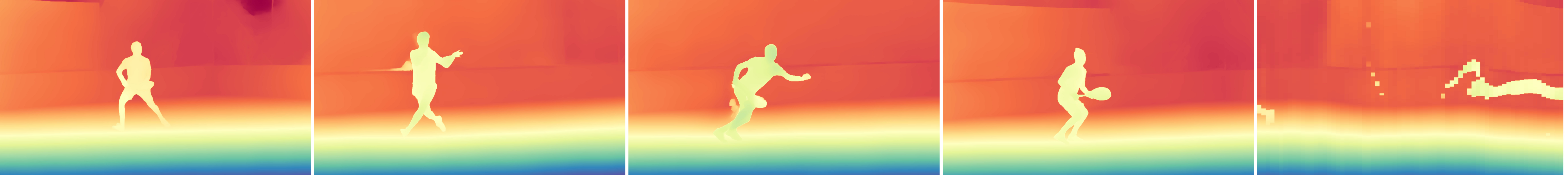} \\
  \end{tabular}
  \caption{The flickering artifacts of FlashDepth and FlashDepth-s~\cite{flashdepth} are clearly visible. Our oVDA produces the sharpest and temporally most stable results.}
  \label{fig:qualitative_tennis}
\end{figure*}
}
{
\setlength{\tabcolsep}{2pt}
\begin{figure*}[h]
  \centering
  \begin{tabular}{@{}m{0.03\linewidth} m{0.97\linewidth}@{}}
    \centering\rotatebox{90}{RGB Video} &
    \includegraphics[width=\linewidth]{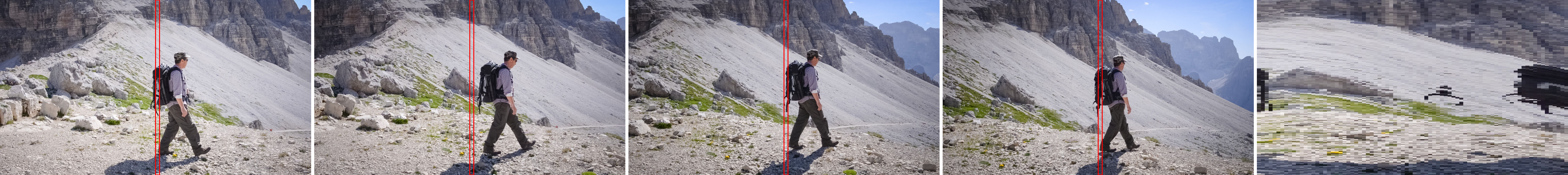} \\[-2pt]
    
    \centering\rotatebox{90}{NVDS} &
    \includegraphics[width=\linewidth]{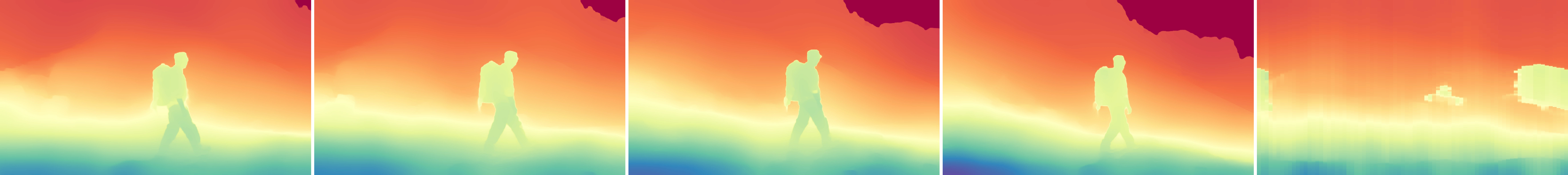} \\[-2pt]

    \centering\rotatebox{90}{ChronoDepth} &
    \includegraphics[width=\linewidth]{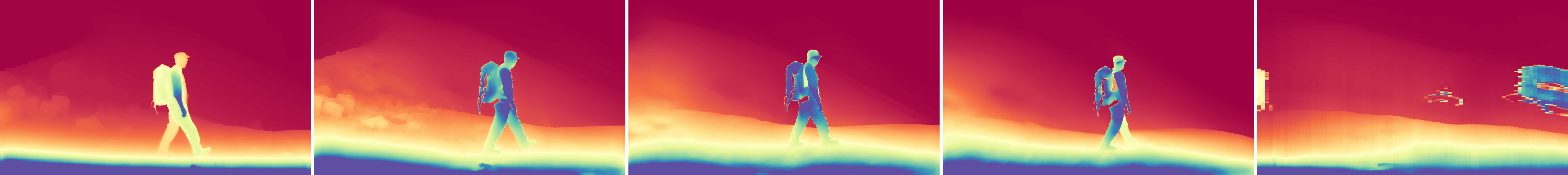} \\[-2pt]
    
    \centering\rotatebox{90}{CUT3R} &
    \includegraphics[width=\linewidth]{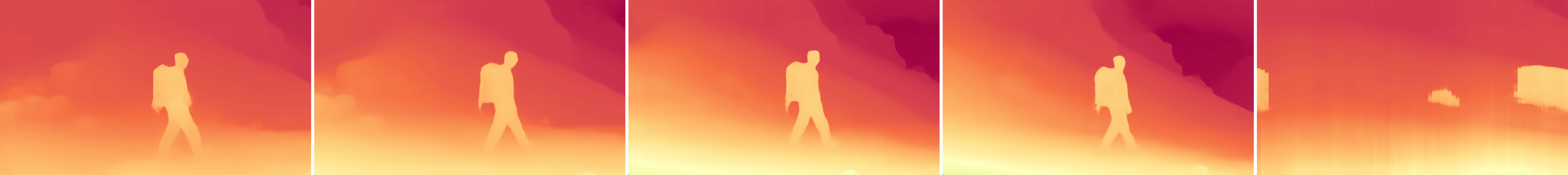} \\[-2pt]
    
    \centering\rotatebox{90}{FlashDepth} &
    \includegraphics[width=\linewidth]{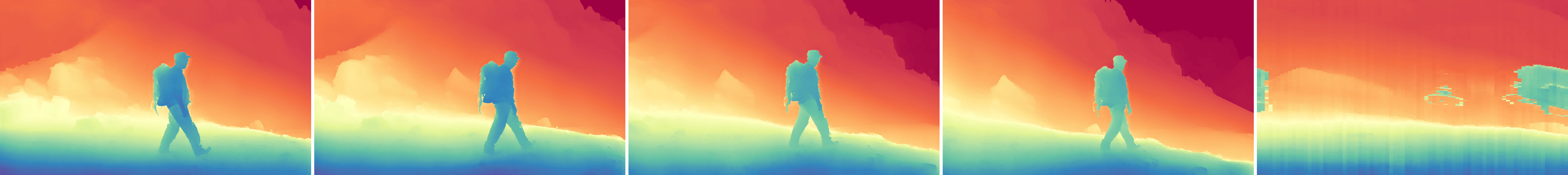} \\[-2pt]
  	
  	\centering\rotatebox{90}{FlashDepth-s} &
    \includegraphics[width=\linewidth]{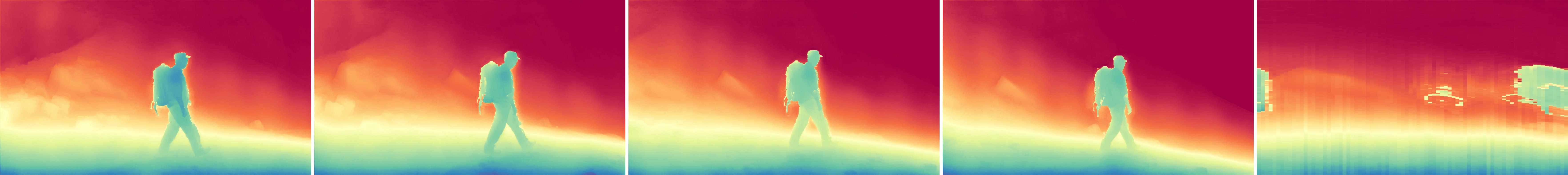} \\[-2pt]
    
    \centering\rotatebox{90}{Ours} &
    \includegraphics[width=\linewidth]{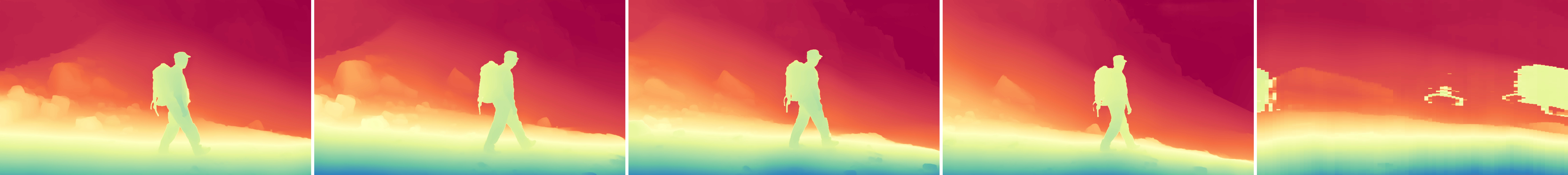} \\
  \end{tabular}
  \caption{Our oVDA generates sharp and temporally consistent predictions.}
  \label{fig:qualitative_hike}
\end{figure*}
}
{
\setlength{\tabcolsep}{2pt}
\begin{figure*}[h]
  \centering
  \begin{tabular}{@{}m{0.03\linewidth} m{0.97\linewidth}@{}}
    \centering\rotatebox{90}{RGB Video} &
    \includegraphics[width=\linewidth]{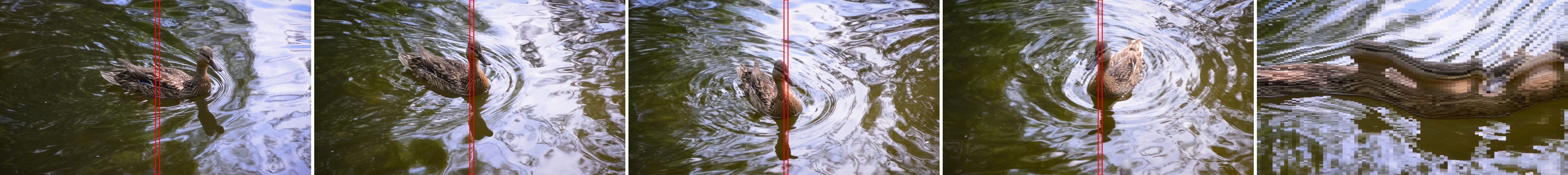} \\[-2pt]
    
    \centering\rotatebox{90}{NVDS} &
    \includegraphics[width=\linewidth]{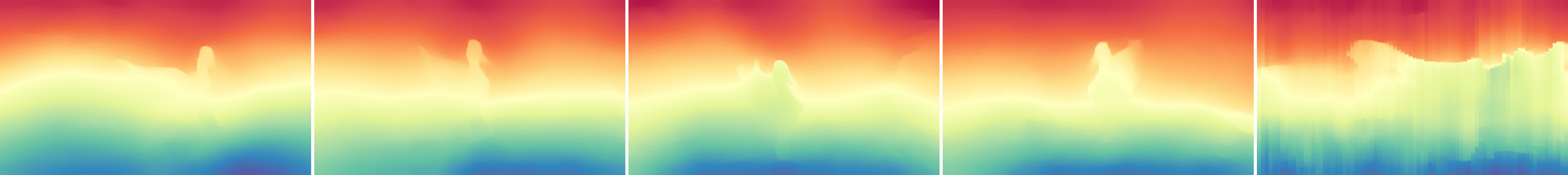} \\[-2pt]

    \centering\rotatebox{90}{ChronoDepth} &
    \includegraphics[width=\linewidth]{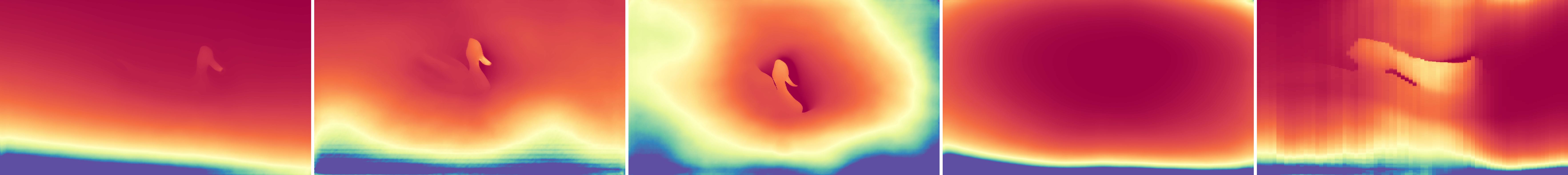} \\[-2pt]
    
    \centering\rotatebox{90}{CUT3R} &
    \includegraphics[width=\linewidth]{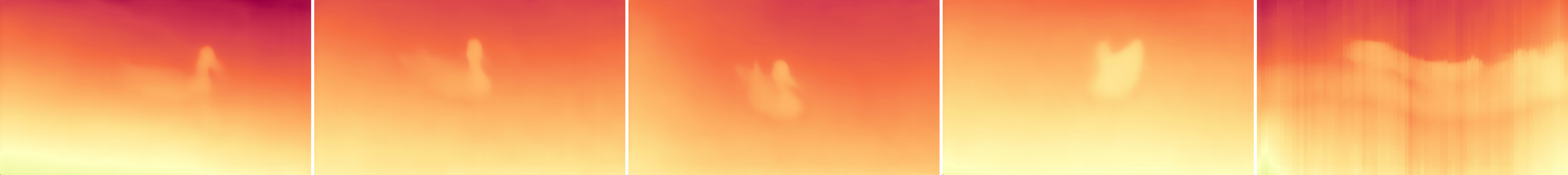} \\[-2pt]
    
    \centering\rotatebox{90}{FlashDepth} &
    \includegraphics[width=\linewidth]{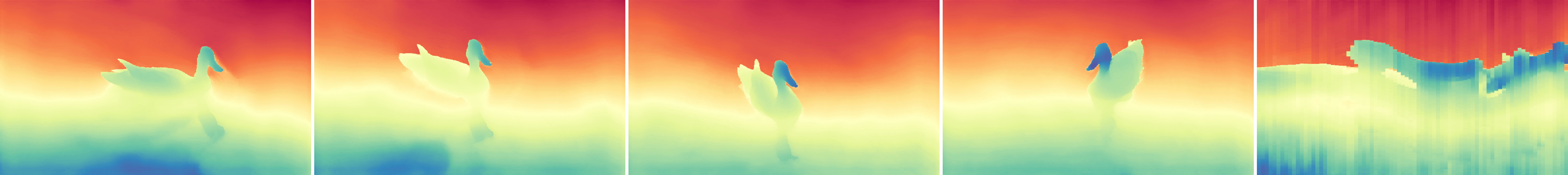} \\[-2pt]
  	
  	\centering\rotatebox{90}{FlashDepth-s} &
    \includegraphics[width=\linewidth]{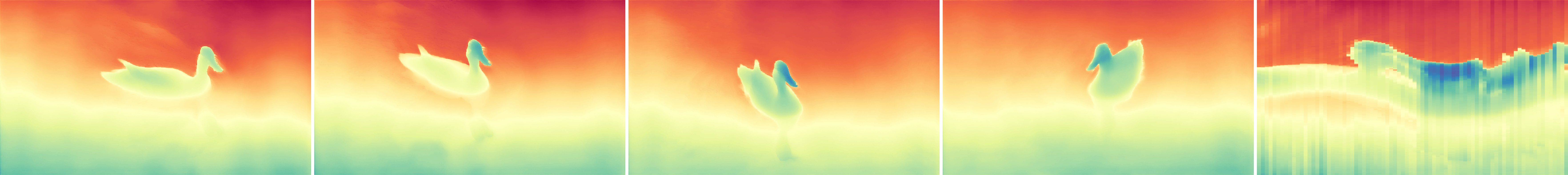} \\[-2pt]
    
    \centering\rotatebox{90}{Ours} &
    \includegraphics[width=\linewidth]{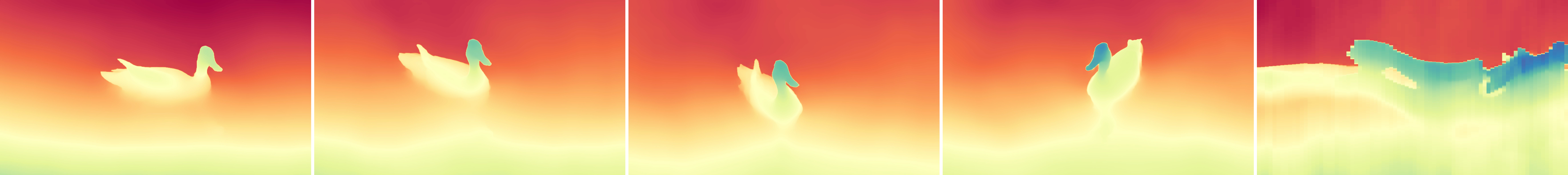} \\
  \end{tabular}
  \caption{The predictions of ChronoDepth~\cite{ChronoDepth} collapse towards the end of this scene, as ChronoDepth was evaluated in online mode. In contrast, our oVDA produces temporally consistent and stable predictions.}
  \label{fig:qualitative_duck}
\end{figure*}
}

\end{document}